\documentclass{article}

\usepackage[nonatbib,main,final]{neurips_2026}

\usepackage[utf8]{inputenc} 
\usepackage[T1]{fontenc}    
\usepackage{hyperref}       
\usepackage{url}            
\usepackage{booktabs}       
\usepackage{amsfonts}       
\usepackage{nicefrac}       
\usepackage{microtype}      
\usepackage{xcolor}         

\usepackage{silence}
\usepackage{graphicx}
\usepackage{amsmath, amssymb, amsthm, mathtools}
\usepackage{booktabs}
\usepackage{multirow}
\usepackage{colortbl}
\usepackage{arydshln}
\usepackage[most]{tcolorbox}
\tcbuselibrary{breakable}
\usepackage{enumitem}
\usepackage{cite}
\usepackage{algorithm}
\usepackage{algorithmic}
\usepackage{wrapfig}

\definecolor{lightorange}{RGB}{246 180 143}
\definecolor{lightred}{RGB}{173, 23, 89}
\definecolor{maroon}{RGB}{122 0 25}
\definecolor{gold}{RGB}{255 204 51}
\hypersetup{
    colorlinks=false,
    pdfborder={0 0 1},
    citebordercolor=lightorange,
    linkbordercolor=white,
    urlbordercolor=white
}

\DeclareMathOperator*{\argmin}{arg\,min}
\usepackage{graphicx,amsfonts,amscd,amssymb,bm,url,color,latexsym,bbm,amsthm,amsmath}
\usepackage{physics}

\usepackage{amsmath, amssymb}
\usepackage{tikz}
\usepackage{xcolor}
\usepackage{graphicx}
\usepackage{subcaption}

\allowdisplaybreaks
\usepackage[capitalize,nameinlink]{cleveref}
\crefname{equation}{Eq.}{Eqs.}
\crefname{figure}{Fig.}{Figs.}
\crefname{table}{Tab.}{Tabs.}
\crefname{section}{Sec.}{Secs.}
\crefname{algorithm}{Alg.}{Algs.}
\crefname{theorem}{Thm.}{Thms.}
\crefname{lemma}{Lem.}{Lems.}
\crefname{corollary}{Cor.}{Cors.}

\newcommand{\mc}{\mathcal}

\newcommand{\bb}{\mathbb}
\newcommand{\set}[1]{\left\{ #1 \right\}}

\newcommand{\eps}{\varepsilon}

\newcommand{\brac}{\bqty}

\newcommand{\wh}{\widehat}

\numberwithin{equation}{section}

\NewCommandCopy{\originalcref}{\cref}
\NewCommandCopy{\originalCref}{\Cref}
\RenewDocumentCommand{\cref}{s m}{%
    \textcolor{lightorange!40!black}{\IfBooleanTF{#1}{\originalcref*{#2}}{\originalcref{#2}}}%
}
\RenewDocumentCommand{\Cref}{s m}{%
    \textcolor{lightorange!40!black}{\IfBooleanTF{#1}{\originalCref*{#2}}{\originalCref{#2}}}%
}

\title{Aligning Language Models with Selective Prediction}

\author{%
  Gaoxiang Luo$^1$, Yifan Wu$^2$, Sinian Zhang$^3$, Aryan Deshwal$^1$, Ju Sun$^1$ \\
    $^1$Department of Computer Science and Engineering\\
    $^2$Bioinformatics and Computational Biology Program\\
    $^3$Division of Biostatistics and Health Data Science\\ 
    University of Minnesota Twin Cities \\
  \texttt{\{luo00042,wu001810,zhan9381,adeshwal,jusun\}@umn.edu} \\
  \\
  Project website: \url{https://sun-umn.github.io/RLSR}
}

\begin{document}

\maketitle

\begin{abstract}
Large language models (LLMs) are increasingly deployed as critical decision-making components in high-stakes real-world AI systems, rendering LLM reliability a foremost practical concern. In this paper, we focus on enhancing LLM reliability through selective prediction (SP), a strategy that allows an LLM to only predict for inputs where it is likely to be correct (i.e., \emph{coverage}) and hence reduce the error rate (i.e., \emph{risk}) on that portion of inputs---flagging the remaining inputs for future human discretion. In other words, SP improves LLM reliability by balancing the risk-coverage trade-off and enabling seamless human-AI collaboration. To integrate SP into LLMs, we focus on the LLM post-training alignment stage and propose to align LLMs with SP performance metrics, in contrast with existing LLM alignment methods that focus primarily on correctness or calibration metrics. Specifically, we propose a novel alignment framework, Reinforcement Learning for Selection Reward (RLSR), which targets the area under the risk-coverage curve (AURC)---a popular SP performance metric---as its alignment objective. RLSR achieves substantially better risk-coverage trade-off compared to multiple alignment baselines on both in-domain and out-of-domain tasks. 
\end{abstract}

\section{Introduction}\label{sec:intro}

In high-stakes applications such as finance, healthcare and law~\cite{llm4lawsurvey,aitrader,llm4clininal}, LLMs are increasingly used to make decisions. In these settings, the cost of making a mistake significantly outweighs that of deferring the case to a human. Thus, it is reasonable to allow and encourage the LLM to abstain when its answer is likely to be wrong. This is the goal of \textbf{selective prediction (SP)}, where the LLM answers only a selected subset of queries that it is likely to get right, to keep its error rate acceptably low on that subset. This paradigm stands in stark contrast to standard LLM training and evaluation, which requires LLMs to answer regardless of their capabilities, causing confident mistakes and hallucinations~\cite{whyllmhallucinate}. In this work, we propose aligning LLMs directly with SP in a principled manner. 

\begin{figure}[t]
    \centering
    \includegraphics[width=0.98\textwidth]{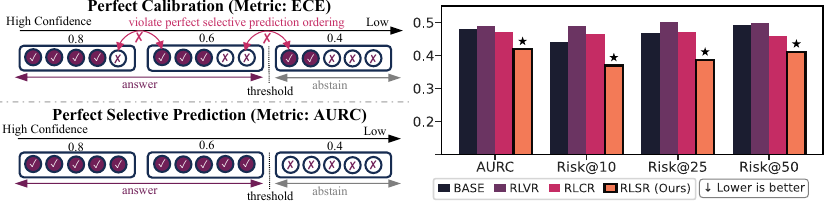}
    \caption{\textbf{(Left)} Perfect confidence calibration does not translate to perfect selective prediction, and vice versa; \textbf{(Right)} Our RLSR aligns LLMs with selective prediction (SP) and substantially improves upon competing methods in terms of SP metrics: AURC and $\operatorname{Risk@}k$.} 
    \label{fig:teaser}
    \vspace{-1em}  
\end{figure}

\vspace{-0.5em}
\paragraph{Confidence calibration (CC) is not sufficient or necessary for selective prediction (SP).}
SP is often assumed, \emph{explicitly or implicitly}, to be the logical next step after confidence calibration (CC)~\cite{guo2017calibration,nixon2019measuring,fischcalibrated,liang2024selective,zhang2026ensemble}. Indeed, assume  that the prediction confidence is perfectly calibrated (i.e., aligned with the true posterior probability $p(y|x)$); then a selection threshold at $0.8$ ensures that the error rate on the selected subset is less than $1-0.8 = 0.2$. While this is technically elegant and convenient, CC and SP are still disparate: CC stresses that the assigned confidence be consistent with the likelihood of correctness, whereas SP favors that correct predictions are ranked ahead of wrong ones—in terms of assigned confidence. As shown in \cref{fig:teaser} (Left), \emph{perfect CC might not lead to perfect SP}, as any threshold we draw can still lead to accepted wrong answers and rejected right answers; conversely, as shown in \cref{fig:teaser} (Right), \emph{perfect SP can be based on ill-calibrated confidence}, so long as all right answers are ranked ahead of wrong ones. Their difference is also reflected by the different evaluation metrics used in practice: CC is often measured by the \emph{expected calibration error} (ECE) or variants, whereas SP by the \emph{area under the risk-coverage curve} (AURC) or variants. Thus, \emph{if the goal is SP, it is technically sound to pursue it directly instead of through CC}. 

\vspace{-0.5em}
\paragraph{Aligning LLMs with selective prediction.}
In this paper, we propose aligning pretrained LLMs directly with SP. To this end, we adopt the popular Reinforcement Learning with Verifiable Rewards (RLVR)~\cite{shao2024deepseekmath} alignment framework, which uses verifiable rewards to improve LLM accuracy and reasoning. The recent work, Reinforcement Learning with Calibration Rewards (RLCR)~\cite{rlcr}, proposes augmenting the correctness reward in RLVR with CC-aware rewards (e.g., the Brier score) to promote LLM accuracy and CC. However, it does not necessarily optimize SP for LLMs, due to the misalignment of CC and SP discussed above. Thus, we propose replacing the correctness reward with SP-aware rewards. Toward this, the AURC, a standard SP metric that summarizes the whole spectrum of risk-coverage tradeoffs, seems a natural choice. However, we face two implementation and performance challenges: (1) the canonical AURC only penalizes errors but does not reward correctness, impairing learning; (2) AURC is a population-level metric that depends on the global ranking of sample confidence, not friendly for mini-batch stochastic gradient methods. 

\vspace{-0.5em}
\paragraph{Our contributions.} 
In this paper, we propose \textbf{Reinforcement Learning for Selection Reward (RLSR)} to directly align pretrained LLMs with SP. To tackle the challenges identified above, we develop a lifted AURC objective, and a batch training strategy based on a judicious choice of AURC reformulation. Our key contributions are:  
\begin{enumerate}[nosep,leftmargin=*]
\item We propose aligning pretrained LLMs with selective prediction (SP) to enhance their reliability for deployment. To the best of our knowledge, this is \emph{the first work} proposing this; 

\item We introduce RLSR, a new LLM alignment framework with an SP-aware reward. We base our learning algorithm on the Group Relative Policy Optimization (GRPO) framework~\cite{shao2024deepseekmath} and develop a novel lifted AURC objective and a batch training strategy to tackle the key technical challenges in optimizing the AURC metric (\cref{sec:our_method}); 

\item We empirically demonstrate that RLSR outperforms strong baselines, including accuracy-oriented RLVR and calibration-oriented RLCR, achieving strong SP performance on multiple in-domain and out-of-domain tasks (\cref{fig:teaser} (Right), \cref{sec:main_exps,sec:sp_exps}). 
\end{enumerate}

\section{Method}\label{sec:method}

\subsection{Technical Background}\label{sec:technical_background}
\paragraph{LLM alignment via RL.}
Let $\mc X$ and $\mc Y$ be the prompt and response spaces, respectively, and $\mc D$ be the data distribution on $\mc X \times \mc Y$. Given a stochastic LLM $\pi_{\theta}$ that maps prompts to responses, a typical LLM alignment step is to improve LLM responses by optimizing 
\begin{equation} \label{eq:alignment-obj}
\max\nolimits_\theta J(\theta) \doteq \mathbb{E}_{(x, y) \sim \mc D} \bb E_{\wh{y} \sim \pi_\theta(\cdot|x)} R(y, \wh{y}), 
\end{equation}
where $\pi_\theta(\cdot|x)$ denotes the distribution on the response space $\mc Y$ conditioned on a prompt $x$---not a single response because $\pi_\theta$ is stochastic, and $R: \mc Y \times \mc Y \to \mathbb{R}$ is a chosen \textbf{reward function}. In other words, LLM alignment is to finetune the LLM parameters $\theta$ by maximizing the expected reward. To maximize using gradient ascent, one can apply the celebrated log-derivative trick, which is popularly used in policy gradient methods for reinforcement learning (RL), to derive the expected gradient as (see \cref{sec:rl-basics} for details)
\begin{align}
    \nabla_\theta J(\theta) = \mathbb{E}_{(x, y)} \mathbb{E}_{\wh{y}} \left[ \nabla_\theta \log \pi_\theta(\wh{y}|x) R(y, \wh{y}) \right], 
\end{align}
where we omit the $\mc D$ and $\pi_{\theta}(\cdot | x)$ under expectation signs to simplify the notation. In practice, the expected gradient is approximated by mini-batch gradient: one estimates the gradient by sampling a batch of $(x, y)$ pairs from $\mc D$, and for each $x$ sampling a batch of $\wh{y}$, i.e., rollouts, from $\pi_{\theta}(\cdot| x)$. 

In practice, various techniques have been proposed to improve the performance of the above framework. We follow the popular \textbf{GRPO} algorithm \cite{shao2024deepseekmath}, which is essentially the classic \textbf{proximal policy optimization (PPO)} in RL with a twist on the advantage function. PPO changes the optimization objective in \cref{eq:alignment-obj} in three aspects: \textbf{(1) rewards into group-relative advantages} by replacing $R(y, \wh{y})$ with prompt-level advantage $A(y, \wh{y})$ that gives more advantages to ``breakthrough'' responses that stand out from the rest; \textbf{(2) per-iteration surrogation} by approximating the objective by  
\begin{align}\label{eq:ppo-ratio}
    J(\theta; \theta_{\mathrm{old}}) \doteq \mathbb{E}_{(x, y)} \mathbb{E}_{\wh{y} \sim \pi_{\theta_{\mathrm{old}}}(\cdot|x)}  \tfrac{\pi_{\theta}(\wh{y}|x)}{\pi_{\theta_{\mathrm{old}}}(\wh{y}|x)} A(y, \wh{y}),  
\end{align}
where $\theta_{\mathrm{old}}$ denotes the model parameters from the previous iteration. This helps decouple response sampling from an old model $\pi_{\theta_{\mathrm{old}}}$ and gradient estimation for the current model $\pi_\theta$; \textbf{(3) range clipping on density ratios} to stabilize training by modifying \cref{eq:ppo-ratio} as 
\begin{align}  
    r_{x, \wh{y}}  \to \min(1+\eps, r_{x, \wh{y}}) \bb I\set{A(y, \wh{y}) > 0} +  \max(1-\eps, r_{x, \wh{y}}) (1 - \bb I\set{A(y, \wh{y}) > 0})
\end{align}
where $r_{x, \wh{y}} \doteq {\pi_{\theta}(\wh{y}|x)}/{\pi_{\theta_{\mathrm{old}}}(\wh{y}|x)}$, i.e., avoid excessively positive and negative advantages so that the update to $\theta$ becomes less biased toward any particular $(x, \wh{y})$ pair.  GRPO, as a variant of PPO, works with the special advantage function 
\begin{equation}
A_{\mathrm{GRPO}}(y, \wh{y}_i) = \brac{R(y, \wh{y}_i) - \text{mean}\set{R(y, \wh{y}_j)}_{j=1}^G}/{\text{std}\set{R(y, \wh{y}_j)}_{j=1}^G}, 
\end{equation}
where $j$ indexes the sampled responses (i.e., rollouts) for the same prompt. This advantage function effectively performs prompt-level zero-mean-unit-variance normalization to the rewards. The gradient of the modified objective in GRPO can be derived using the log-derivative trick, analogous to the process for \cref{eq:alignment-obj}. 

\paragraph{RLVR \& RLCR.} 
RLVR has recently emerged as the gold-standard LLM alignment framework for mathematical and coding tasks~\cite{deepseekr1,wen2025reinforcement}. It typically assumes an external programmatic verifier to judge the correctness of the response and gives a binary reward 
\begin{align} \label{eq:rlvr-reward}
R_{\text{RLVR}}(y, \wh{y}) \doteq \mathbb{I}\{y = \wh{y} \}.
\end{align}
After RLVR training, recent reasoning models often return verbalized confidence~\cite{kumaran2026how,design_prompt} alongside their responses, explicitly communicating the model's internal uncertainty to the user (see \cref{app:prompts}).
Although $R_{\text{RLVR}}$ explicitly incentivizes accurate prediction, RLVR-finetuned LLMs are often found to be overconfident, i.e., not well calibrated~\cite{taming_overconfidence}. To promote both correct prediction and calibrated confidence, RLCR~\cite{rlcr} proposes adding a Brier score penalty to the RLVR reward, leading to 
\begin{align} \label{eq:rlcr-reward}
R_{\text{RLCR}}(y, \wh{y}, s) \doteq \mathbb{I}\{y = \wh{y} \} - \left(s- \mathbb{I}\{y = \wh{y}\}\right)^2, 
\end{align}
where $s$ denotes the verbalized confidence returned by the LLM alongside $\wh{y}$. 

\paragraph{Selective prediction (SP).} 
\begin{wrapfigure}{r}{0.46\textwidth}
    \centering
    \vspace{-1em}  
    \includegraphics[width=0.46\textwidth]{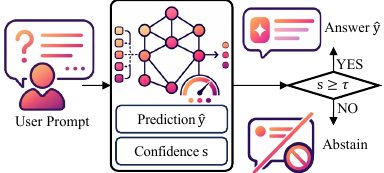}
    \caption{SP for LLMs in deployment.}
    \label{fig:sp_in_action}
    \vspace{-1em}  
\end{wrapfigure}
For a general input-output space $\mc X' \times \mc Y'$ and an associated data distribution $\mc D'$ on $\mc X' \times \mc Y'$, SP augments a predictive model $f: \mc X' \to \mc Y'$ with a binary selector $g: \mc X' \to \set{0, 1}$, so that the final output by the pair $(f, g)$ is: 
\begin{equation} \label{eq:llm_sc_def}
(f, g)(x') =
\begin{cases}
f(x') & \text{if } g(x')=1,\\
\text{abstain} & \text{if } g(x')=0
\end{cases}
\; \forall x' \in \mc X'. 
\end{equation}
The selector $g$ typically takes the form 
\begin{equation} \label{eq:selector}
g_{s, \tau}(x')
\doteq
\mathbb{I}\!\left\{ s(x') \ge \tau \right\}, 
\end{equation}
where $s: \mc X' \to [0, 1]$ is a confidence scorer and $\tau \in [0, 1]$ is a selection threshold. The selector chooses to keep only the most confident predictions based on the confidence scorer $s$ with a confidence cutoff $\tau$. To evaluate the performance of SP, two metrics are of particular interest: the \textbf{coverage} is the portion of predictions \emph{effectively} covered by the selector, i.e., not rejected by the selector 
\begin{equation}
\phi_{s, \tau}(f) 
\doteq
\mathbb{E}_{(x',y')\sim\mathcal{D'}}\!\left[g_{s, \tau}(x')\right]; 
\label{eq:coverage}
\end{equation}
the \textbf{selective risk} is the risk over the covered subpopulation 
\begin{equation}
R_{s, \tau}(f) \doteq
{
\mathbb{E}_{(x',y')\sim\mathcal{D'}}
\!\left[\mathbb{I}\!\left\{f(x') \neq y\right\} g_{s, \tau}(x')\right]
}/{
\phi_{s, \tau}(f) 
}, 
\label{eq:selective_risk}
\end{equation}
where the binary loss $\mathbb{I}\!\left\{f(x') \neq y\right\}$ could be replaced by other appropriate losses. For any fixed $(f, s)$, there is typically a tradeoff between coverage and selective risk determined by $\tau$. So, one can vary $\tau$ over $[0, 1]$ to plot the entire tradeoff curve, called the \textbf{risk-coverage curve}. A popular summary metric for quantifying SP performance is the \textbf{area under the risk-coverage curve} (AURC). For a finite dataset of size $N$, the coverage varies by $1/N$ (assuming that there is no tie in the confidence). So, the empirical AURC is computed as  
\begin{align}
\overline{\mathrm{AURC}} \doteq \frac{1}{N} \sum\nolimits_{i=1}^{N} R_{(i)}, \quad  \text{where} \; R_{(i)} \;&=\; \frac{1}{i} \sum\nolimits_{j=1}^{i} \mathbb{I}\{f(x_{(j)}) \neq y_{(j)}\}
\label{eq:aurc_def}
\end{align}
i.e., $R_{(i)}$ denotes the selective risk computed on the subset of the most confident samples $i$ (the subscript $\cdot_{(j)}$ indicates the sample with the $j$-th highest confidence score). Clearly, many of the indicator terms are repeated across different $R_{(i)}$'s. The equivalent form of \cite{zhou2024novel} collects the repeated indicator terms together so that each sample appears inside the summation exactly once, i.e., as a weighted empirical loss 
\begin{align} \label{eq:aurc_reweighted}
\overline{\mathrm{AURC}}^{\mathrm{w}} = \frac{1}{N} \sum\nolimits_{i=1}^{N} \widehat{\alpha}_i \cdot \mathbb{I}\{f(x_i) \neq y_i\}, \quad \text{with} \; \widehat{\alpha}_i = H_N - H_{N-r_i} = \sum\nolimits_{j=1}^{r_i} \tfrac{1}{N - j + 1}. 
\end{align} 
Here, $r_i$ denotes the ranking of the $i$-th sample in ascending order of confidence, and $H_N$ represents the $N$-th harmonic number. This weighted form is our primary focus below thanks to its conciseness.

\subsection{Our Method: Reinforcement Learning for Selection Reward (RLSR)}\label{sec:our_method}

As discussed in \cref{sec:intro}, although RLVR and RLCR could improve LLM accuracy and perhaps also confidence calibration, they are not sufficient to boost LLM SP performance. Our goal here is to directly align pretrained LLMs with SP based on the GRPO framework described in \cref{sec:technical_background}. To implement GRPO, it is natural to design the reward to be consistent with $\overline{\mathrm{AURC}}^{\mathrm{w}}$ defined in \cref{eq:aurc_reweighted}. For SP, we choose to work with \textbf{verbalized confidence} as an exemplar confidence, following RLCR~\cite{rlcr} and due to its simplicity and recent popularity~\cite{design_prompt,kumaran2026how,survey_confidence,survey_uncertainty}.   

\vspace{-0.5em}
\paragraph{Two challenges with AURC.}
The negative version of $\overline{\mathrm{AURC}}^{\mathrm{w}}$ in \cref{eq:aurc_reweighted} can serve as a natural reward at the population-level. However, two fundamental challenges arise when we try to integrate it into GRPO: 
\begin{enumerate}[nosep,leftmargin=*]
    \item[(1)] \textbf{Sparse rewards.} \quad  $\overline{\mathrm{AURC}}^{\mathrm{w}}$ is a weighted sum of binary indicators of the form $\mathbb{I}\{f(x_i) \neq y_i\}$, which produces sparse one-sided learning signals with two deficiencies: 
    \textbf{(i) No signals for correct predictions.} Since $\overline{\mathrm{AURC}}^{\mathrm{w}}$ only penalizes errors and does not assign learning signals to correct predictions, this could slow down learning or even impair it; \textbf{(ii) No margin enforcement.} Since correct predictions are not rewarded regardless of their confidence levels, a confident correct prediction and a lucky guess are treated equally. This would allow the confidence margin between correct and wrong predictions to be arbitrarily close---the alignment process has no incentives to promote it, which can in turn hurt generalization performance; 
    \vspace{0.5em}
    \item[(2)] \textbf{Rewards based on ranking.} \quad The definition of $\overline{\mathrm{AURC}}^{\mathrm{w}}$ in \cref{eq:aurc_reweighted} depends on $\widehat{\alpha_i}$, which depends on the ranking of the $i$-th sample over the entire dataset. This means that $\overline{\mathrm{AURC}}^{\mathrm{w}}$ is a population-level metric that depends on the global ranking of all samples. However, mini-batch stochastic gradient descent (SGD) applied to the GRPO framework clearly relies on mini-batch rewards in each step, which cannot be decoupled from the global ranking. But global reranking per iteration step would defeat the purpose of mini-batch SGD in avoiding computation over the whole dataset. 
\end{enumerate}

\vspace{-0.5em}
\paragraph{Lifted AURC to address the first challenge.} 
We introduce a reformulation called \emph{Lifted AURC}. The idea is to shift the binary indicator $\mathbb{I}\{\wh{y_i} \neq y_i\} \in \{0,1\}$ to a signed variable $2\mathbb{I}\{\wh{y_i} \neq y_i\} \in \{0,1\}- 1 \in \{-1, +1\}$ directly on \cref{eq:aurc_reweighted}, resulting in
\begin{align}
\overline{\mathrm{AURC}}^{\mathrm{w}}_{\mathrm{lift}} &= \frac{1}{N} \sum\nolimits_{i=1}^N \widehat{\alpha}_i \cdot \bigl(2 \mathbb{I}\{\wh{y_i} \neq y_i\} - 1\bigr).  \label{eq:lifted_reweighted}
\end{align} 
Since $ \frac{1}{N}\sum\nolimits_{i=1}^N \widehat{\alpha}_i \cdot \bigl(2 \mathbb{I}\{\wh{y_i} \neq y_i\} - 1\bigr) = \frac{2}{N}\sum\nolimits_{i=1}^N \widehat{\alpha}_i \cdot \mathbb{I}\{\wh{y_i} \neq y_i\} -  \frac{1}{N} \sum\nolimits_{i=1}^N \widehat{\alpha}_i = \frac{2}{N}\sum\nolimits_{i=1}^N \widehat{\alpha}_i \cdot \mathbb{I}\{\wh{y_i} \neq y_i\} -  1$, we have $\overline{\mathrm{AURC}}^{\mathrm{w}}_{\mathrm{lift}}= 2\overline{\mathrm{AURC}}^{\mathrm{w}} - 1$. So, minimizing $\overline{\mathrm{AURC}}^{\mathrm{w}}_{\mathrm{lift}}$ is equivalent to minimizing $\overline{\mathrm{AURC}}^{\mathrm{w}}$.\footnote{The equivalence is in the sense that they have the same set of global solutions. But due to the significant overparameterization of LLMs, the global solution is unlikely to be unique and they may have different levels of generalization. } This lifted version $\overline{\mathrm{AURC}}^{\mathrm{w}}_{\mathrm{lift}}$ provides two-sided rewards that actively reward correct predictions and penalize incorrect ones. The opposite signs of the nonzero rewards for correct and wrong predictions, respectively, create a ``push-pull'' dynamic that drives correct and wrong predictions apart in their confidence rankings, thereby improving the confidence margin. This margin enforcement is critical for SP, which relies on a sufficient separation of confidence between correct and wrong predictions. 
\begin{figure}[!htbp]
    \centering
    \includegraphics[width=0.48\textwidth]{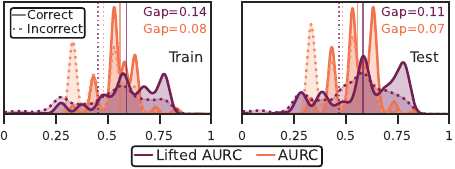}
    \includegraphics[width=0.48\textwidth]{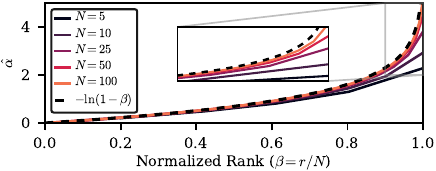}
    \caption{\textbf{(Left)} Normalized confidence distributions of correct and incorrect predictions. Empirically, lifted AURC as a reward achieves larger separation (margin) between correct and incorrect predictions on both train and test sets (see \cref{app:margin_analysis}); \textbf{(Right)} Behavior of the rank-based weights $\widehat{\alpha}$. }
    \label{fig:conf_dist}
\end{figure}
In \cref{fig:conf_dist} (Left), we empirically confirm the margin benefit of using the lifted AURC as a reward, compared to the plain AURC. Consistent with $\overline{\mathrm{AURC}}^{\mathrm{w}}_{\mathrm{lift}}$, we define the per-sample RLSR reward as 
\begin{align}
R_{\text{RLSR}}(y_i, \widehat{y_i}, s_i) = \begin{cases} +\widehat{\alpha}_i & \text{if } \widehat{y}_i = y_i \\ -\widehat{\alpha}_i & \text{if } \widehat{y}_i \neq y_i \end{cases}.
\label{eq:final_reward}
\end{align}
\begin{wrapfigure}{l}{0.5\textwidth}
    \centering
    \vspace{-14pt}
    \includegraphics[width=1.0\linewidth]{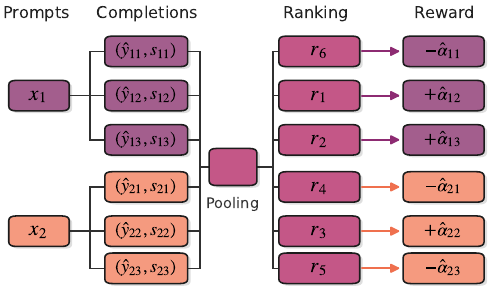}
    \caption{\textbf{Illustration of implementation.} We pool rollouts from $B$ inputs, each generating $G$ samples, into an effective batch of size $B \times G$ (e.g., $B=2$, $G=3$). Rewards are computed based on ranking over the effective batch. The subsequent relative advantage computation is identical to standard GRPO within each input group.}
    \label{fig:implementation}
    \vspace{-1em}
\end{wrapfigure}
To confirm that this definition of the per-sample reward makes intuitive sense, in \cref{fig:conf_dist} (Right), we plot the weights $\widehat{\alpha}$ against the normalized rank $\beta = r/N$ (in ascending order of confidence) for varying batch sizes $N \in \{5, \dots, 100\}$. The empirical weights converge to the theoretical curve $-\log(1-\beta)$~\cite{zhou2024novel}. Moreover, this weighting scheme places much more emphasis on high-confidence predictions than on low-confidence ones (the asymptotic slope of the weight curve moving from low-confidence to high-confidence is $1/(1-\beta)$). This is sensible, as high-confidence ones are likely to persist in the selected subset when one gradually increases the selection threshold $\tau$, than low-confidence ones that tend to be rejected early. Thus, attending more to the high-confidence ones helps improve the overall SP performance, aligned with our goal. 

\vspace{-0.5em}
\paragraph{Batch training to address the second challenge.}
Although $R_{\text{RLSR}}$ provides a reasonable reward per-sample, computing the weight $\widehat{\alpha}_i$ still requires ranking across all samples. We address this by the following batch training strategy (see \cref{fig:implementation}). For each mini-batch of $B$ samples, we independently generate $G$ rollouts for each sample and then pool all these $B \times G$ rollouts into an effective batch. We then compute ranks $r_i$ and weights $\widehat{\alpha}_i$ across this pooled set, approximating the population-level AURC. The batch-level ranking step introduces negligible overhead and remains compatible with standard GRPO throughout. The pseudocode of the entire algorithm can be found in \cref{alg:rlsr_grpo}. 

\begin{wrapfigure}{r}{0.6\textwidth}
\vspace{-24pt}
\begin{minipage}{0.6\textwidth}
\begin{algorithm}[H]
\small
\caption{Our algorithm: GRPO with $R_{\text{RLSR}}$ in \cref{eq:final_reward}}
\label{alg:rlsr_grpo}
\begin{algorithmic}[1]
\REQUIRE LLM $\pi_\theta$, data $\mathcal{D}$,
  generations $G$, clipping $\eps$,
  regex parser $\mathcal{P}$ for answer and verbalized confidence
\FOR{step $= 1, \ldots, T$}
  \STATE Sample batch $\{(x_b, y_b)\}_{b=1}^B$ from $\mathcal{D}$
  \STATE $\pi_{\theta_{\mathrm{old}}} \leftarrow \pi_\theta$
  \FOR{$b = 1, \ldots, B$}
    \STATE Sample $G$ outputs from $\pi_{\theta_{\mathrm{old}}}(\cdot \mid x_b)$
    \STATE $(\widehat{y}_{b,g},\, s_{b,g}) \leftarrow \mathcal{P}(\pi_{\theta_{\mathrm{old}}}(\cdot \mid x_b))$ for $g = 1,\ldots,G$
  \ENDFOR
  \STATE \textit{// Pool $n {=} B {\times} G$ rollouts for batch-level ranking}
  \STATE Sort all $n$ samples ascendingly;
    assign $r_i \in \{1,\ldots,n\}$
  \STATE $\widehat{\alpha}_i \leftarrow H_n - H_{n-r_i}$ \qquad
    \COMMENT{where $H_k \doteq \sum_{j=1}^{k} 1/j$}
  \STATE $R_i \leftarrow
    \begin{cases}
      +\widehat{\alpha}_i & \widehat{y}_i = y_i \\
      -\widehat{\alpha}_i & \widehat{y}_i \neq y_i
    \end{cases}$
  \STATE \textit{// Group-relative advantages (GRPO)}
  \FOR{$b = 1, \ldots, B$}
    \STATE $A_{b,g} \leftarrow
      \left(R_{b,g} - \mathrm{mean}_g(R_{b,\cdot})\right) / \mathrm{std}_g(R_{b,\cdot})$
  \ENDFOR
  \STATE \textit{// Clipped policy update (per-token)}
  \STATE $\rho_{i,t} \leftarrow
    \pi_\theta(o_{i,t} \mid x, o_{i,<t}) \,/\,
    \pi_{\theta_{\mathrm{old}}}(o_{i,t} \mid x, o_{i,<t})$
  \STATE $\mathcal{L} \leftarrow
    -\mathbb{E}_{i,t}\!\big[
      \min\!\big(
        \rho_{i,t} A_i,\,
        \mathrm{clip}(\rho_{i,t}, 1{\pm}\eps) A_i
      \big)
    \big]$
  \STATE Update $\pi_\theta$ by $\nabla_\theta \mathcal{L}$
\ENDFOR
\STATE \textbf{return} $\pi_\theta$
\end{algorithmic}
\end{algorithm}
\end{minipage}
\vspace{-32pt}
\end{wrapfigure}

\vspace{-0.5em}
\paragraph{Comparison with RLVR.} 
When rollouts for an input are all correct or wrong, RLVR assigns the same reward to all, yielding zero advantage and no learning signal. However, this \emph{sparse reward} issue is less likely for RLSR because it differentiates rollouts by confidence ranks, providing a unique reward to each rollout (if no ties or ties are broken by averaging $\widehat{\alpha}_i$). This allows the model to separate low-confidence lucky guesses from high-confidence predictions and penalize confident errors more heavily than uncertain mistakes. 

\section{Related Work}\label{sec:related_work}
\vspace{-0.5em}
\paragraph{Confidence scorers for LLMs.} 
In the SP literature, confidence scorers are often derived from prescribed rules (e.g., these based on output logits) or learned confidence predictors~\cite{liang2024selective}. For LLMs, rule-based confidence scorers proposed in the literature are numerous, but the community has not converged to a standard~\cite{selectivepredictiongap}. Recent work~\cite{aspire,selectllm} has also developed learned confidence scorers for LLMs. Our RLSR framework is compatible with both types of confidence scorers, as our reward only requires a confidence scorer that ranks the samples alongside a correctness verifier.

\vspace{-0.5em}
\paragraph{Direct learning with AURC.} 
In machine learning, learning with the AURC objective is still rare. Recently, \cite{franc2023optimal,zhou2024novel} has explored this for classification problems, but their focuses are mostly on the effective computation and approximation of AURC for learning and optimization purposes; \cite{zhang2026ensemble} learns the ensemble of post-hoc scorer given a set of base scorers, toward optimizing the AURC loss. Our method in \cref{sec:our_method} hinges on the key equivalent definition of AURC as weighted empirical risk (i.e., \cref{eq:aurc_reweighted}) developed in \cite{zhou2024novel}. However, the two technical challenges we address in \cref{sec:our_method} are unique to the GRPO framework (and RL problems in general) and our RLSR algorithm, and therefore not considered in~\cite{franc2023optimal,zhou2024novel}.  

We include more discussion of remoter related work in \cref{app:more_related_work}. 

\vspace{-1pt}
\section{Experiments}
To verify the efficacy of our RLSR alignment framework, we benchmark RLSR against RLVR and RLCR in terms of risk-coverage tradeoff on both in-domain and out-of-domain tasks in \cref{sec:main_exps}. We further evaluate these alignment frameworks under a risk-controlled deployment setting in high-stakes domains in \cref{sec:sp_exps}. 

\subsection{Benchmarking in risk-coverage tradeoff}\label{sec:main_exps}
\paragraph{Models and datasets.}
We consider both Qwen2.5-7B~\cite{qwen2025qwen25technicalreport} and Llama-3.1-8B~\cite{grattafiori2024llama3herdmodels} as our base LLM models. We perform alignment training on two datasets: HotPotQA-Modified~\cite{hotpotqa} and BigMath~\cite{bigmath}. HotPotQA-Modified is a multi-hop question answering dataset that tests reasoning under incomplete or distracting evidence, while BigMath is a multi-step mathematical reasoning dataset where uncertainty accumulates across steps. \emph{To diversify the evaluation sets and cover different uncertainty types other than those in the training datasets, we include $6$ additional datasets:} SimpleQA and TriviaQA~\cite{simpleqa,triviaqa} probe overconfidence on rare factual knowledge; GPQA, Math500, and GSM8K~\cite{gpqa,math500,gsm8k} assess multi-step reasoning; and CommonsenseQA~\cite{commonsenseqa} examines ambiguous, implicit reasoning scenarios. To construct the in-domain (ID) and out-of-domain (OOD) evaluation sets, we organize the datasets as follows: for HotPotQA-aligned models, ID evaluation only uses HotPotQAs, and OOD evaluation includes the remaining $6$ additional datasets; for BigMath-aligned models, Math500, GSM8K, and BigMath are for ID evaluation, while the rest are for OOD evaluation. 

\vspace{-0.5em}
\paragraph{Competing methods, confidence scorers, and evaluation metrics.}
\textbf{Competing methods}. The key baseline methods we include are (1) BASE, no alignment, (2) RLVR, the standard correctness-only alignment, and (3) RLCR, the recent correctness+calibration alignment. We also compare against non-alignment alternatives such as supervised finetuning in \cref{app:sft_baselines};  \textbf{Confidence scorers}. To elicit verbalized confidence and reasoning about uncertainty, we include a format reward that encourages the model to output an analysis and a confidence score after the chain-of-thought answer (see \cref{app:training_hyperparams}). Since our method is agnostic to the choice of the confidence scorer, we also experiment with other post-hoc confidence scorers for the selector; RLSR attains the best or tied-best AURC under every scorer (see \cref{app:conf_ablation}); \textbf{Evaluation metrics}. Our primary metric is the standard SP metric AURC, which directly measures the holistic risk-coverage tradeoff. Alongside AURC, we also report the accuracy (i.e., $1-$risk) at the 10\%, 25\%, and 50\% coverage levels. High-stakes applications typically require operating in the low-risk (i.e., high-accuracy) regime, which corresponds to low coverage. For completeness, we also report ECE, the standard calibration metric. However, note that ECE is not a direct measure of SP performance and is not optimized by RLSR. 

\vspace{-0.5em}
\paragraph{Training and inference details.}
For each prompt, we generate responses with temperature $T=0.7$ and use an effective batch size of $B \times G = 1536$ samples. We then rank the samples inside the effective batch by their confidence scores to compute the RLSR reward (\cref{alg:rlsr_grpo}). To elicit structured outputs, we adapt the RLCR system prompts for HotPotQA and BigMath, which induce \texttt{<think>}, \texttt{<answer>}, \texttt{<analysis>}, and \texttt{<confidence>} tags. The maximum response length is $1536$ tokens for HotPotQA and $4096$ tokens for BigMath. All competing methods share identical training configurations and operate under equal inference budgets for fair comparison; full hyperparameter details are in \cref{app:training}. Following~\cite{outcomerl}, we omit the std normalization when calculating the GRPO advantage; an ablation study confirms the difference is minor (HotPotQA ID AURC: $0.46$ with std normalization vs.\ $0.44$ without)).
In evaluation, all results use greedy decoding ($T=0$) for reproducibility; statistical significance over 5 seeds at $T=0.3$ is reported in \cref{app:stat_significance}, and robustness to the choice of LLM judge in \cref{app:judge}. 

\begin{table*}[htbp]
\centering
\vspace{-6pt}
\caption{Comparison of SP performance on in-domain (ID) (2 for HotPotQA and 3 for BigMath) and out-of-domain (OOD) evaluation datasets (6 for HotPotQA and 5 for BigMath) with the Qwen2.5-7B model. \colorbox{lightorange!15}{Orange} highlights the best result per metric within each training setting. RLSR also outperforms all baselines on the Llama-3.1-8B model; see  \cref{tab:llama_results}.} 
\setlength{\tabcolsep}{2.5pt}
\small
\begin{tabular}{@{}l@{\hspace{6pt}}ccccc@{\hspace{8pt}}ccccc@{}}
\toprule
 & \multicolumn{5}{c}{\textbf{In-Domain}} & \multicolumn{5}{c}{\textbf{Out-of-Domain}} \\
\cmidrule(lr){2-6} \cmidrule(l){7-11}
Method & AURC$\downarrow$ & ECE$\downarrow$ & Acc@10$\uparrow$ & Acc@25$\uparrow$ & Acc@50$\uparrow$ & AURC$\downarrow$ & ECE$\downarrow$ & Acc@10$\uparrow$ & Acc@25$\uparrow$ & Acc@50$\uparrow$ \\
\midrule
\multicolumn{11}{l}{\textit{HotPotQA Training}} \\
\midrule
BASE & 0.60 & 0.57 & 46.0\% & 40.6\% & 37.8\% & 0.44 & 0.40 & 59.1\% & 57.4\% & 54.9\% \\
RLVR & 0.56 & 0.56 & 45.5\% & 42.0\% & 41.8\% & 0.47 & 0.46 & 52.6\% & 52.9\% & 53.1\% \\
RLCR & 0.51 & \cellcolor{lightorange!15}0.05 & 53.5\% & 49.0\% & 50.4\% & 0.46 & \cellcolor{lightorange!15}0.23 & 53.5\% & 54.3\% & 55.3\% \\
\hdashline
\textbf{RLSR} & \cellcolor{lightorange!15}0.44 & 0.24 & \cellcolor{lightorange!15}66.0\% & \cellcolor{lightorange!15}63.2\% & \cellcolor{lightorange!15}57.2\% & \cellcolor{lightorange!15}0.41 & 0.28 & \cellcolor{lightorange!15}62.0\% & \cellcolor{lightorange!15}60.7\% & \cellcolor{lightorange!15}59.5\% \\
\midrule
\multicolumn{11}{l}{\textit{BigMath Training}} \\
\midrule
BASE & 0.41 & 0.40 & 58.9\% & 61.5\% & 57.9\% & 0.49 & 0.44 & 54.7\% & 51.1\% & 49.9\% \\
RLVR & 0.28 & 0.25 & 72.2\% & 70.7\% & 71.5\% & 0.53 & 0.49 & 46.4\% & 43.9\% & 46.7\% \\
RLCR & 0.26 & \cellcolor{lightorange!15}0.12 & 76.7\% & 75.5\% & 73.6\% & 0.47 & \cellcolor{lightorange!15}0.21 & 57.3\% & 53.8\% & 52.1\% \\
\hdashline
\textbf{RLSR} & \cellcolor{lightorange!15}0.24 & 0.18 & \cellcolor{lightorange!15}78.8\% & \cellcolor{lightorange!15}78.4\% & \cellcolor{lightorange!15}76.4\% & \cellcolor{lightorange!15}0.45 & 0.34 & \cellcolor{lightorange!15}58.5\% & \cellcolor{lightorange!15}56.3\% & \cellcolor{lightorange!15}54.2\% \\
\bottomrule
\end{tabular}
\label{tab:results}
\vspace{-5pt}
\end{table*}

\begin{figure*}[htbp]
    \centering
    \vspace{-8pt}
    \includegraphics[width=1.0\linewidth]{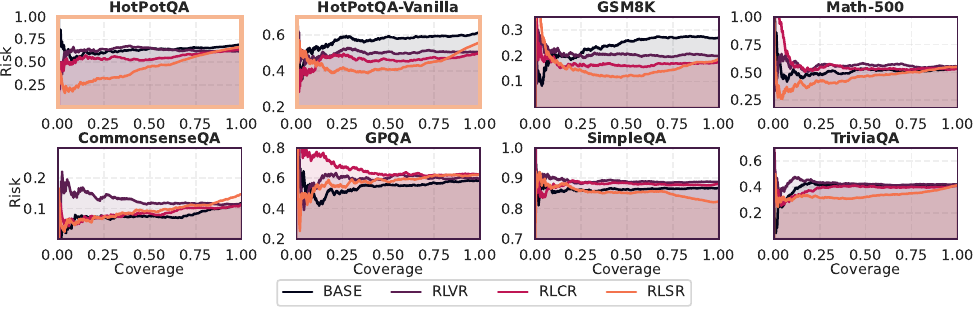}
    \caption{The risk-coverage curves of the Qwen2.5-7B model aligned on HotPotQA. The in-domain datasets are framed in {\tiny \fcolorbox{orange}{white}{\textcolor{orange}{orange}}} while the remaining are out-of-domain datasets.}
    \label{fig:aurc_hotpot}
    \vspace{-5pt}
\end{figure*}
\begin{figure*}[htbp]
    \centering
    \vspace{-5pt}
    \includegraphics[width=1.0\linewidth]{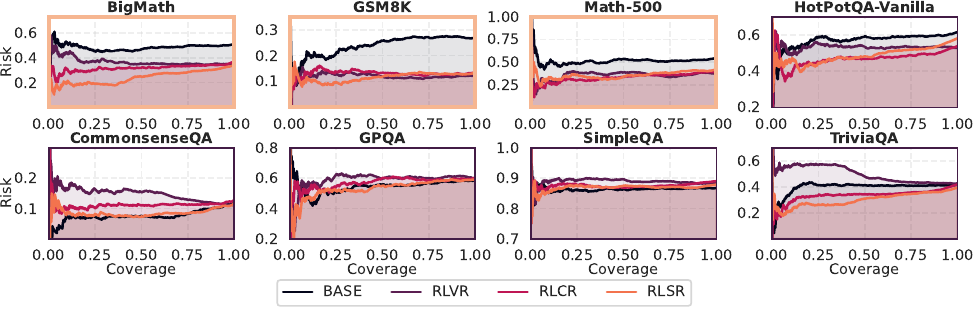}
    \caption{The risk-coverage curves of the Qwen2.5-7B model aligned on BigMath. The in-domain datasets are framed in {\tiny \fcolorbox{orange}{white}{\textcolor{orange}{orange}}} while the remaining are out-of-domain datasets.}
    \label{fig:aurc_math}
    \vspace{-1em}
\end{figure*}
\vspace{-0.5em}
\paragraph{RLSR's superior in-domain performance.}
From \cref{tab:results}, it is evident that RLSR consistently outperforms RLVR and RLCR in terms of both AURC and accuracy-at-coverage across all three coverage levels for ID evaluation. For example, for models trained on HotPotQA, RLSR achieves an AURC of $0.44$ compared to $0.56$ for RLVR and $0.51$ for RLCR; and at $10\%$ coverage, RLSR reaches $66.0\%$ accuracy versus $45.5\%$ for RLVR and $53.5\%$ for RLCR. Models trained on BigMath exhibit similar gains across all metrics. Notably, RLCR achieves the best ECE but lags behind RLSR in AURC, whereas RLSR has higher ECE than RLCR but the best AURC. This provides direct empirical evidence for the misalignment between calibration and SP discussed in \cref{sec:intro} and \cref{fig:teaser} (Left): better calibration does not imply better SP, and vice versa. So, RLSR is the better choice when selection is the goal. 

\vspace{-0.5em}
\paragraph{RLSR's SP capability transferable to novel domains.} 
RLSR's SP performance remains strong on OOD evaluation, as shown in  \cref{tab:results}. It uniformly improves over all baseline methods, and particularly by large margins on the HotPotQA group. Surprisingly, RLVR always degrades the OOD SP performance relative to the base method without any alignment, and RLCR degrades the OOD SP performance in HotPotQA, but improves the performance in BigMath. We suspect that such performance degradation is partially due to overfitting the ID distribution during the alignment process, which makes the model less robust to distribution shifts. Notably, RLSR's SP performance advantage remains consistent despite the distribution shifts. The advantage also extends to code generation, a task type absent from both training sets: on MBPP~\cite{mbpp}, RLSR attains the best AURC, Acc@10, and Acc@25 for both training domains (\cref{app:mbpp}).

\begin{wraptable}{l}{0.45\textwidth}
\centering
\vspace{-14pt}
\caption{Controlled confidence gap analysis on HotPotQA ID (Qwen2.5-7B). We restrict to cases 
where all 3 methods predict correctly or incorrectly, controlling accuracy as a confounder. $\Delta c = \mathbb{E}[c \vert \text{correct}] - \mathbb{E}[c \vert \text{incorrect}]$ measures how well each method separates correct from incorrect predictions by confidence.} 
\renewcommand{\arraystretch}{1.15}
\setlength{\tabcolsep}{4pt}
\small
\begin{tabular}{@{}lccc@{}}
\toprule
Method & $\mathbb{E}[c \vert \text{correct}]$ & $\mathbb{E}[c \vert \text{incorrect}]$ & $\Delta c$ $\uparrow$ \\
\midrule
RLVR & 1.00 & 0.98 & 0.02 \\
RLCR & 0.45 & 0.39 & 0.06 \\
\hdashline
\textbf{RLSR} & \textbf{0.89} & \textbf{0.52} & \textbf{0.37} \\
\bottomrule
\end{tabular}
\label{tab:delta_c}
\vspace{-10pt}
\end{wraptable}

\vspace{-0.5em}
\paragraph{Discussion.} 
To understand where RLSR's performance gains originate, we examine the detailed behavior of the risk-coverage curves in \cref{fig:aurc_hotpot,fig:aurc_math}. A revealing pattern emerges: at full coverage, RLSR typically achieves comparable (or even slightly higher) risk than RLVR and RLCR, indicating similar overall accuracy. However, as coverage decreases, the RLSR's risk drops substantially faster, yielding a lower AURC. This is expected: the risk at full coverage is the population risk by definition, whereas RLSR targets AURC rather than the population risk. The gains are not limited to very low coverage: RLSR achieves the best Acc@50 in all four settings of \cref{tab:results}. This behavior directly confirms the benefit of our alignment objective: by optimizing a lifted AURC reward, RLSR learns to assign higher confidence to correct predictions and lower confidence to incorrect ones (see \cref{tab:delta_c} for empirical separation), precisely the ranking needed for effective selection. In contrast, RLCR shows a more gradual decline, and RLVR's curve often remains flat or even increases at low coverage, implying inconsistent confidence assignment for totally ignoring confidence in RLVR. 

\vspace{-0.5em}

\paragraph{Ablation study.} Since RLSR approximates the global ranking through batch ranking, we study the effect of the ranking pool size $B\times G$ on SP performance. Fixing $G=32$ and training Qwen2.5-7B on HotPotQA with $B=48, 32, 16$, the ID AURC is $0.44, 0.44, 0.45$ and the OOD AURC is $0.41, 0.47, 0.48$, respectively. The ID performance is stable, while the OOD performance degrades with smaller batches. The latter degradation is concentrated on the two math OOD benchmarks (\cref{app:batch_ablation}). Halving the group size to $G=16$, which halves the rollout cost, degrades AURC only slightly (ID $0.46$, OOD $0.42$ at $B=48$), still outperforming RLVR and RLCR at the full cost. 

\begin{wrapfigure}{r}{0.55\textwidth}
    \centering
    \vspace{-1em}
    \includegraphics[width=1.0\linewidth]{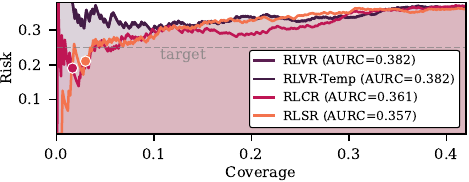}
    \caption{The risk-coverage curves of the three alignment methods on the MedQA validation set. The dashed line indicates the $25\%$ target risk (i.e., $75\%$ accuracy), and the dots mark the actual operating point of each method---not exactly on $25\%$ risk level due to threshold selection on the finite validation set.}
    \label{fig:medqa_aurc}
    \vspace{-2em}
\end{wrapfigure}
\subsection{Benchmarking in Risk-Controlled Deployment}\label{sec:sp_exps}
To show the effectiveness of RLSR in high-stakes deployment, we apply the models trained on HotPotQA to MedQA~\cite{medqa}, a medical question-answering benchmark derived from various professional medical board and licensing examinations. This represents a challenging OOD setting where risk control is critical. 

\vspace{-0.5em}
\paragraph{Experimental setup.} 
Since all alignment methods on MedQA achieve full-coverage accuracy between $57\%$ and $59\%$ (see \cref{fig:medqa_aurc}), we set a target accuracy at $75\%$, much higher than the full-coverage accuracy. We focus on this low-coverage, low-risk regime with a controlled risk level to emulate practical high-stakes deployment scenarios. To achieve the target accuracy, we find the cutoff threshold $\tau$ for each method on a validation set with samples $1,272$, and then apply the threshold to the test set with samples $1,273$.

\begin{wraptable}{r}{0.55\textwidth}
\centering
\vspace{-13pt}
\caption{SP performance of the three methods on MedQA (target accuracy: $75\%$). Temperature scaling (\texttt{RLVR-Temp}) compresses RLVR's confidence range to $[0.38, 0.62]$, so $\tau=0.62$ is the maximum achievable threshold; $(\cdot)^\dagger$ highlights methods not able to achieve the target accuracy.}
\setlength{\tabcolsep}{3pt}
\small
\resizebox{\linewidth}{!}{%
\begin{tabular}{@{}lccccc@{}}
\toprule
\textbf{Method} & $\tau$ & \textbf{Val. Cov.} & \textbf{Val. Acc.} & \textbf{Test Cov.} & \textbf{Test Acc.} \\
\midrule
RLVR$^\dagger$ & 1.0 & -- & 66.6\% & -- & -- \\
RLVR-Temp$^\dagger$ & 0.62 & -- & 66.6\% & -- & -- \\
RLCR & 0.95 & 1.7\% & 81.0\% & 1.4\% & 72.2\% \\
\hdashline
\textbf{RLSR} & 1.0 & 3.0\% & 78.9\% & 3.0\% & \textbf{78.9\%} \\
\bottomrule
\end{tabular}
}
\label{tab:medqa_deployment}
\vspace{-13pt}
\end{wraptable}

\vspace{-0.5em}
\paragraph{Results and discussion.}
As shown in \cref{tab:medqa_deployment} and \cref{fig:medqa_aurc}, RLVR fails to achieve the target accuracy $75\%$ throughout the coverage range, due to their overly high and ill-calibrated confidence regardless of the prediction correctness---the very issue that RLCR aims to address~\cite{rlcr,taming_overconfidence}. Post-hoc temperature scaling~\cite{guo2017calibration,calibrating_verbalized} (see \cref{sec:temp-scaling-detail}) over RLVR (RLVR-Temp) does not improve or even change the SP performance, as it simply performs monotonic rescaling of the confidence score and hence preserves confidence ranking. This underscores the importance of optimizing the confidence scorers in RLCR and RLSR. RLSR's advantage is most pronounced in the low-coverage, low-risk regime (see \cref{fig:medqa_aurc}): RLSR's risk-coverage curve is not only almost uniformly lower than that of RLCR's, but reaches very low risk---RLCR's reaches a significant risk floor ($\sim 15\%$) first and then bounces back as coverage level decreases. For error control (see \cref{tab:medqa_deployment}), only RLSR and RLCR achieve the $75\%$ target accuracy on the validation set but have very different generalization behaviors: while RLCR's test accuracy drops below the target ($81.0\%\; \text{validation} \rightarrow 72.2\%\; \text{test}$), RLSR has the same test accuracy and coverage as the validation ones. Overall, RLSR obtains simultaneous higher accuracy and coverage than RLCR's by large margins on the test set, demonstrating RLSR's strong error-controlled SP performance. The absolute coverage of all methods for this stringent target is low. Whether it is acceptable depends on the cost of errors relative to that of abstention in the application. For a fixed target risk, absolute coverage can be increased with a stronger LLM and confidence scorer, or with a risk-controlled alignment objective (\cref{sec:limitations}).

\section{Conclusion}
For an LLM to be reliable in high-stakes deployment, it must not only answer correctly but also know when to answer. Our central message is that calibration is neither sufficient nor necessary for this goal: selective prediction (SP) depends on whether correct answers are ranked ahead of incorrect ones by confidence, which calibration does not guarantee. Hence, if the goal is SP, LLMs should be aligned with SP directly. Our alignment method, RLSR, does so by optimizing the area under the risk-coverage curve (AURC) as a reward. In our evaluation across in-domain and out-of-domain benchmarks, RLSR achieves significantly better risk-coverage tradeoff than accuracy-oriented (RLVR) and calibration-oriented (RLCR) alignment; and in a risk-controlled simulated medical deployment task, it is the only method that meets the target accuracy with reasonable coverage. 

\section*{Limitations and Future Works and Broader Impacts}\label{sec:limitations}
\paragraph{Limitations.}
\textbf{First}, AURC summarizes the entire risk-coverage tradeoff, so optimizing it does not guarantee better coverage at any target risk level, and coverage at a stringent risk could be very low (\cref{sec:sp_exps}). \textbf{Second}, RLSR optimizes the confidence ranking rather than its calibration, so its ECE is higher than that of RLCR. This reflects a practical tension between calibration and selection. In fact, the only possibility they are perfectly aligned is when all correct predictions get confidence $1$ and all incorrect confidence $0$---unlikely attainable in practice. \textbf{Third}, RLSR approximates the population-level ranking in AURC by ranking within a batch, which introduces finite-sample approximation noise. The OOD performance degrades with smaller batches (\cref{sec:main_exps}), and we do not provide a theoretical characterization of this noise. \textbf{Fourth}, like other alignment methods that rely on a verifier, RLSR requires automatically verifiable correctness, so open-ended generation and multi-turn dialog remain outside its current scope. 

\paragraph{Future works.}
When the target risk is known, one could formulate risk-controlled coverage maximization as the alignment objective. Beyond global calibration, which is misaligned with SP, one could pursue calibration on the selected subset~\cite{fisch2022calibrated}. A finite-sample analysis of batch ranking would clarify how batch size affects the approximation of AURC. Extending RLSR to open-ended and multi-turn tasks would be extremely interesting. 

\paragraph{Broader Impacts.}
This work focuses on the refusal capacity of LLMs so that the retained subset can be more accurate. The potential positive societal impacts include reliable decision making in high-stakes domains because the users can expect that the returned answer comes from the retained subset, and thus the predictive accuracy is close to the target accuracy set for pre-deployment calibration. Just like any optimization methods, if users turn the risk axis in AURC into any misuse metric, this technology can be turned into a selective prediction system for selectively returning the misuse information only. At the end of the day, it is really up to the intent of the users.

\begin{ack}
This work was supported in part by a UMN DSI-MnDRIVE
PhD Graduate Assistantship, the National Institutes of Health (NIH) under award No. R01CA287413, R01NS131314, R01MH138929, U01FD008720, the National Science Foundation (NSF) under award No. 2435911 and 2541023. The content is solely the responsibility of the authors and does not necessarily represent the official views of the National Institutes of Health. The authors acknowledge the Minnesota Supercomputing Institute (MSI) at the University of Minnesota for providing resources that contributed to the research results reported within this paper. 
\end{ack}

\clearpage
\bibliography{reference}
\bibliographystyle{IEEEtran}


\clearpage
\appendix
\section{Derivations \& Proofs}\label{app:proof}

\subsection[Gradient of the LLM alignment objective]{Gradient of the LLM alignment objective in \texorpdfstring{\cref{eq:alignment-obj}}{Equation}}
\label{sec:rl-basics}
First of all, 
\begin{align}
    \nabla_{\theta} J(\theta) 
    = \nabla_{\theta} \mathbb{E}_{(x, y) \sim \mc D} \bb E_{\wh{y} \sim \pi_\theta(\cdot|x)} R(y, \wh{y}) 
    = \mathbb{E}_{(x, y) \sim \mc D} \nabla_{\theta} \bb E_{\wh{y} \sim \pi_\theta(\cdot|x)} R(y, \wh{y}),  
\end{align} 
so we can focus on deriving $\nabla_{\theta} \bb E_{\wh{y} \sim \pi_\theta(\cdot|x)} R(y, \wh{y})$. Now 
\begin{align}
    \nabla_{\theta} \bb E_{\wh{y} \sim \pi_\theta(\cdot|x)} R(y, \wh{y})
    & = \nabla_{\theta} \sum\nolimits_{\wh{y}} \pi_\theta(\wh{y}|x) R(y, \wh{y})  \\
    & = \sum\nolimits_{\wh{y}} \nabla_{\theta} \pi_\theta(\wh{y}|x) R(y, \wh{y})  \\
    & = \sum\nolimits_{\wh{y}} \pi_\theta(\wh{y}|x) \nabla_\theta \log \pi_\theta(\wh{y}|x) R(y, \wh{y})  \quad (\text{log-derivative trick}) \\
    & = \bb E_{\wh{y} \sim \pi_\theta(\cdot|x)} \brac{\nabla_\theta \log \pi_\theta(\wh{y}|x) R(y, \wh{y})},  
\end{align}
as claimed. 

\section{More Related Work}\label{app:more_related_work}
\vspace{-0.5em}
\paragraph{Selective algorithms for LLMs.} 
While our paper focuses on LLM alignment and optimizing LLMs for SP, several other lines of orthogonal work focusing on other aspects of LLM deployment also contain selection components. For example, selective answering~\cite{selective_answering} deals with selecting among inherently ambiguous questions; post-abstention~\cite{post_abstention} addresses the question of what to do after abstention, i.e., refine selective prediction; conformal LLM~\cite{conformal_language_modeling} adapts conformal prediction, which outputs uncertainty sets, for LLMs; and guided/constrained decoding~\cite{liang2024controllable} modifies the LLM autoregressive generation process by selecting and filtering tokens to ensure its output respects prescribed constraints. 

\begin{wrapfigure}{r}{0.5\textwidth}
    \vspace{-1em}
    \centering
    \includegraphics[width=1.0\linewidth]{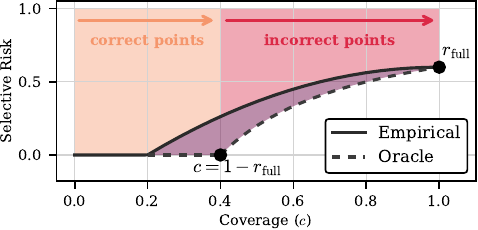}
    \caption{The dashed curve is the oracle under which coverage levels left of $c=1-r_{\text{full}}$ (\textcolor{lightorange}{orange}) accept all correct predictions first and rank incorrect predictions last (\textcolor{lightred}{red}). $r_{\text{full}}$ means full-coverage risk. The \textcolor{violet}{purple} shaded area visualizes the SP gap.} 
    \label{fig:sp_gap}
    \vspace{-1em}
\end{wrapfigure}

\vspace{-0.5em}
\paragraph{Confidence calibration (CC) for LLMs.} 
Overconfidence in LLMs trained with RL with human feedback (RLHF) is widely recognized~\cite{taming_overconfidence,rlcr}. So, CC has been a dominant line of research around it~\cite{tian2023just}, evidenced by a wide range of post-hoc CC methods (e.g., temperature scaling) shown to improve calibration metrics such as expected calibration error (ECE)~\cite{temp_scaling,calibrating_by_fidelity,luo-deshwal-2025-com,calibrate_w_consistency}. But, our focus of this paper is selective prediction, as discussed in \cref{sec:intro}.

\vspace{-0.5em}
\paragraph{Selective prediction (SP) gap.} 
In SP, the ideal selector should mimic the \textit{perfect-ordering oracle}, which ranks all correct predictions ahead of all incorrect ones in terms of confidence, thus minimizing the AURC. However, practical selectors typically fall short of this performance ceiling. \cite{selectivepredictiongap} formalizes this shortfall as the \textit{SP gap} (see \cref{fig:sp_gap}) and decomposes it into four major sources: Bayes noise, approximation limit, statistical noise, and ranking error. For LLMs, while the approximation limit and statistical noise can be collectively reduced with larger model size and data volume during training, the major obstacle to closing the gap comes from the ranking error, which is determined by the confidence scorer. The goal of our work is to improve the LLMs capabilities as simultaneous predictors and confidence scorers by aligning them with SP, thus improving the ranking error. \cite{selectivepredictiongap} also points out that many post-hoc calibration methods do not improve the ranking error because they are monotone transformations of the original scores and hence leave the ranking of samples intact. 

\section{Experiment Details}\label{app:exps_details}
\subsection{Prompt Templates}\label{app:prompts}
We use structured prompts that guide the LLM to generate both answers and confidence scores. Following \cite{rlcr}, we prompt the LLM to produce outputs with four components: (1) a thinking trace enclosed in \texttt{<think>} tags, (2) a final answer in \texttt{<answer>} tags, (3) an uncertainty analysis in \texttt{<analysis>} tags, and (4) a numerical confidence score in \texttt{<confidence>} tags:
\begin{tcolorbox}[colback=black!5!white,colframe=black!75!black,title=Example Response]
\footnotesize
\texttt{<think>} The problem is asking for the person who established the Journal of Libertarian Studies as well as [...] but their connection to the establishment of the Journal of Libertarian Studies is not clear.\texttt{</think>} \\
\texttt{<answer>}\textbf{Murray Newton Rothbard}\texttt{</answer>} \\
\texttt{<analysis>} One uncertainty arises from the fact that while [...], the information about him writing over 20 books on political theory is less direct.
It would be better if [...].\texttt{</analysis>} \\
\texttt{<confidence>}\textbf{0.85}\texttt{</confidence>}
\end{tcolorbox}
We use two variants of prompts. For Big-Math Digits, we use the \emph{Simple Prompt}; for HotPotQA and HotPotQA-Vanilla, we use the \emph{Long Prompt}, which provides additional guidelines for uncertainty analysis. 
\begin{tcolorbox}[colback=black!5!white,colframe=black!75!black,title=System Prompt (Simple)]
\label{box:tabc_prompt}
\footnotesize
A conversation between User and Assistant. The user asks a question, and the Assistant solves it. The Assistant first thinks about the reasoning process in the mind, provides the user with the final answer, then analyzes its confidence about the solution and provides the user with its confidence level. The confidence level is a number between 0 and 1 (inclusive) enclosed within \texttt{<confidence>}\texttt{</confidence>} tags. The final answer is enclosed between \texttt{<answer>}\texttt{</answer>} tags. The analysis about confidence and uncertainty is enclosed within \texttt{<analysis>}\texttt{</analysis>} tags. The Assistant should reason about its confidence in the solution and its uncertainty in the solution within these tags. The final format that must be followed is: \texttt{<think>} reasoning process here \texttt{</think>}\texttt{<answer>} final answer here \texttt{</answer>}\texttt{<analysis>} analysis about confidence and uncertainty here \texttt{</analysis>}\texttt{<confidence>} confidence level here (number between 0 and 1) \texttt{</confidence>}
\end{tcolorbox}

\begin{tcolorbox}[breakable,  toprule at break=0mm, 
  bottomrule at break=0mm, colback=black!5!white,colframe=black!75!black,title=System Prompt (Long)]
\label{box:tabc_long_prompt}
\footnotesize
A conversation between User and Assistant. The user asks a question, and the Assistant solves it. The assistant first thinks about the reasoning process in the mind, provides the user with the final answer, then analyzes its confidence about the solution and then provides the user with its confidence level. The confidence level is a number between 0 and 1 (inclusive) enclosed within \texttt{<confidence>}\texttt{</confidence>} tags. The final answer is enclosed between \texttt{<answer>}\texttt{</answer>} tags. The analysis about confidence and uncertainty is enclosed within \texttt{<analysis>}\texttt{</analysis>} tags. The assistant should reason about its confidence in the solution and its uncertainty in the solution within these tags. Here are some guidelines for the analysis:
\begin{enumerate}
    \item Your task is to point out things where the model could be wrong in its thinking, or things where there might be ambiguity in the solution steps, or in the reasoning process itself.
    \item You should not suggest ways of fixing the response, your job is only to reason about uncertainties.
    \item For some questions, the response might be correct. In these cases, it is also okay to have only a small number of uncertainties and then explicitly say that I am unable to spot more uncertainties.
    \item Uncertainties might be different from errors. For example, uncertainties may arise from ambiguities in the question, or from the application of a particular lemma/proof.
    \item If there are alternate potential approaches that may lead to different answers, you should mention them.
    \item List out plausible uncertainties, do not make generic statements, be as specific about uncertainties as possible.
    \item Enclose this uncertainty analysis within \texttt{<analysis>}\texttt{</analysis>} tags.
\end{enumerate}
The final format that must be followed is: \texttt{<think>} reasoning process here \texttt{</think>}\texttt{<answer>} final answer here \texttt{</answer>}\texttt{<analysis>} analysis about confidence and uncertainty here \texttt{</analysis>}\texttt{<confidence>} confidence level here (number between 0 and 1) \texttt{</confidence>}
\end{tcolorbox}

\subsection{Verifiers}\label{app:verifiers}
Verifiers are used during training to compute rewards and during evaluation to assess correctness. We employ three types of verifiers, following~\cite{rlcr}:
\begin{enumerate}[nosep,leftmargin=*]
    \item \textbf{Exact-Match.} The predicted answer must exactly match the ground truth answer string. This verifier is used for HotPotQA and HotPotQA-Vanilla.
    \item \textbf{Math-Verify.} We use \texttt{math-verify}, a robust mathematical expression evaluation system that checks semantic equivalence of mathematical expressions. This verifier is used for Math-500, GSM8K, and Big-Math Digits.
    \item \textbf{LLM-as-a-Judge.} We use Llama-3.1-8B-Instruct with greedy decoding (temperature=0) as the judge model. The judge is provided with the question, the groundtruth answer, and the predicted answer, and is prompted to respond with ``YES'' or ``NO'' based on correctness. Importantly, we do not condition the judge on thinking traces to avoid potential biases. This verifier is used for TriviaQA, SimpleQA, CommonsenseQA, GPQA, and MedQA. The judge is served locally on fixed hardware with fixed software dependencies, so its output is (almost) deterministic. To rule out systematic errors introduced by a single judge, we also re-judge the out-of-domain generations with two additional judges from different model families and of different sizes (\cref{app:judge}). 
    \item \textbf{Code-Execution.} The generated program is executed against the test cases provided with each problem, and it is judged correct if all test cases pass. This verifier is used for MBPP. 
\end{enumerate}
\begin{tcolorbox}[colback=black!5!white,colframe=black!75!black,title=LLM-as-a-Judge Prompt,label=box:judge_prompt]
\footnotesize
You are given a question, a ground truth answer, and a predicted answer. Determine whether the predicted answer is correct based on the ground truth. Respond with only ``YES'' if the predicted answer is correct, or ``NO'' if it is incorrect.

\textbf{Question:} \{question\}

\textbf{Ground Truth Answer:} \{ground\_truth\}

\textbf{Predicted Answer:} \{predicted\_answer\}

\textbf{Is the predicted answer correct?}
\end{tcolorbox}

\subsection{LLM Response Parser}\label{app:answer_confidence_extract}
All responses use the \texttt{<think>}, \texttt{<answer>}, \texttt{<analysis>}, and \texttt{<confidence>} tags. The answers are extracted from the \texttt{<answer>} tags and confidence scores from the \texttt{<confidence>} tags. For models that do not follow the expected format, we append ``Thinking time ended. My final answer is'' to the output and prompt the model again to extract the answer. Similarly, if a valid confidence score cannot be extracted, we append ``Thinking time ended. My verbalized confidence in my answer as a number between 0 and 1 is equal to'' and prompt the model again. \emph{Note that models trained with format rewards (see \cref{app:training_hyperparams}) rarely trigger these fallbacks}.

\subsection{Training}\label{app:training}

\subsubsection{Training Datasets}\label{app:train_datasets}

\begin{enumerate}[nosep, leftmargin=*]
    \item \textbf{HotPotQA-Modified.} We adopt the HotpotQA-Modified dataset introduced by \cite{rlcr}, which is derived from the original HotPotQA distractor dataset~\cite{hotpotqa}. This modified version systematically varies the availability of supporting evidence by removing zero, one, or both of the key paragraphs required to answer each question. The questions are evenly distributed across these three conditions, and each instance contains 8 paragraphs in total. This construction creates varying levels of information completeness, making it particularly suitable for developing and evaluating uncertainty reasoning capabilities. The training set comprises $20,000$ examples, and correctness is evaluated by exact string matching. HotPotQA is distributed under CC BY-SA 4.0 License and HotPotQA-Modified is under MIT License. The convention of this paper uses HotPotQA for the modified version and HotPotQA-Vanilla for the original dataset.
    \item \textbf{BigMath.} We use BigMath~\cite{bigmath}, a large-scale curated dataset for reinforcement learning that contains over $250,000$ mathematical problems. Following \cite{rlcr}, we retain only problems with LLaMA-8B solve rates between 0\% and 70\% to maintain appropriate difficulty, and further restrict to problems with numerical answers to enable reliable automatic verification---resulting in the subset referred to as BigMath Digits. The final training set contains $15,000$ problems, with correctness evaluated using \texttt{math-verify}. It is distributed under the MIT License. 
\end{enumerate}

\subsubsection{Training Details}\label{app:training_hyperparams}
\paragraph{Base LLM models \& GRPO setup.} 
We use Qwen2.5-7B~\cite{qwen2025qwen25technicalreport} under the Apache 2.0 License and Llama3.1-8B~\cite{grattafiori2024llama3herdmodels} under the Llama 3.1 Community License. Following recent work on RL for reasoning~\cite{deepseekr1,openreasonerzero}, we initialize RL directly from the base model without supervised finetuning and set $\beta = 0$ (no KL regularization). We use GRPO without standard deviation normalization in the advantage calculation~\cite{outcomerl} \emph{(an ablation confirms that the difference is minor: HotPotQA ID AURC 0.46 with std normalization vs.\ 0.44 without)}, along with the BNPO loss function~\cite{bnpo}, which aggregates token-level losses using the number of active tokens in the local training batch. For parameter-efficient training, we apply LoRA~\cite{hu2022lora} with rank $r = 1$, $\alpha = 32$, and no dropout to all linear layers, motivated by \textit{LoRA Without Regret}~\cite{schulman2025lora}.   

\paragraph{Reward functions.} 
The total reward combines a task-specific reward with a format reward:
\begin{equation}
    R_{\text{total}} = R_{\{\text{RLVR}, \text{RLCR}, \text{RLSR}\}} + R_{\text{format}}.
    \label{eq:total_reward}
\end{equation}
The method-specific reward is defined in \cref{eq:rlvr-reward} for RLVR, \cref{eq:rlcr-reward} for RLCR, and \cref{eq:final_reward} for RLSR. The format reward $R_{\text{format}} \in \{0, 1\}$ is a binary indicator: $1$ if the output follows the expected tag structure (\texttt{<think>}, \texttt{<answer>}, \texttt{<analysis>}, \texttt{<confidence>}) and contains a valid confidence score $c \in [0, 1]$, and $0$ otherwise. \emph{Note that the format reward converges to near-perfect values early in the training processes for all methods}. 

\vspace{-0.5em}
\paragraph{Key hyperparameters.} 
We generate $32$ responses per prompt with temperature $0.7$. \textbf{For HotPotQA}, we use a learning rate of $1 \times 10^{-5}$ with a constant schedule and $5\%$ warmup, training for $1$ epoch with an effective batch size of $1536$ ($4$ GPUs $\times$ $3$ per-device batch $\times$ $128$ gradient accumulation steps). The maximum prompt and completion lengths are $3072$ and $1536$ tokens, respectively. \textbf{For BigMath}, we use a learning rate of $5 \times 10^{-5}$ with a linear decay schedule and $20\%$ warmup, training for $0.5$ epochs with an effective batch size of $1536$ ($4$ GPUs $\times$ $6$ per-device batch $\times$ $64$ gradient accumulation steps). The maximum prompt and completion lengths are $1024$ and $4096$ tokens, respectively. All methods (RLVR, RLCR, RLSR) use identical hyperparameters to ensure fair comparison. All training uses the Transformers Reinforcement Learning (TRL) library, with vLLM as an efficient rollout engine~\cite{kwon2023efficient}. Important system optimizations include \texttt{bfloat16} (\texttt{bf16}) precision, FlashAttention-2 (FA2)~\cite{dao2023flashattention2} and Triton kernels shipped with TRL. 

\vspace{-0.5em}
\paragraph{Training time.} Training runs on $4$xNVIDIA H100 GPUs, and each run takes approximately $62$ hours (HotPotQA) or $57$ hours (BigMath). This holds for all three alignment methods, since they differ only in the reward computation, and the ranking step of RLSR takes negligible CPU time. 

\vspace{-0.5em}
\subsection{Evaluation}\label{app:eval}
We follow RLCR~\cite{rlcr} for setting up the evaluation protocol. All models are evaluated with greedy decoding (temperature=$0$) and a maximum token budget of $4096$ (sufficient for all selected evaluation benchmarks). We generate a single response per question and use the same system prompts as during training: the simple prompt (\cref{app:prompts}) for Big-Math Digits and the long prompt for HotPotQA. 

\subsubsection{Evaluation Datasets}\label{app:eval_datasets}

\begin{enumerate}[nosep,leftmargin=*] 
    \item \textbf{HotPotQA.} (MIT License) We use $500$ held-out validation samples from the modified dataset introduced by \cite{rlcr}. Correctness is measured using exact-match. 
    \item \textbf{HotPotQA-Vanilla.} (CC BY-SA 4.0 License) We use $1,000$ validation samples from the original HotPotQA distractor dataset~\cite{hotpotqa}, with $2$ irrelevant paragraphs removed from each question. Thus, each question contains $8$ paragraphs with both supporting paragraphs present. Correctness is measured using exact-match.
    \item \textbf{Math-500.} (MIT License) We use the MATH-500 dataset, a subset of problems from the original MATH dataset~\cite{math500}. Correctness is measured using \texttt{math-verify}.
    \item \textbf{GSM8K.} (MIT License) We use $1,319$ problems from the test set of the Grade School Math 8K dataset~\cite{gsm8k}. Correctness is measured using \texttt{math-verify}. 
    \item \textbf{Big-Math Digits.} (MIT License) We use $1,000$ held-out validation samples from the filtered Big-Math dataset~\cite{bigmath}. Correctness is measured using \texttt{math-verify}. 
    \item \textbf{TriviaQA.} (Apache 2.0 License) We use $2,000$ samples from the TriviaQA~\cite{triviaqa} validation set, specifically the no-context split, to test the factual accuracy. Correctness is measured using LLM-as-a-judge. 
    \item \textbf{SimpleQA.} (MIT License) We use the full dataset consisting of $4,326$ factual questions~\cite{simpleqa}. Correctness is measured using LLM-as-a-judge. 
    \item \textbf{CommonsenseQA.} (MIT License) uses $1,220$ problems from the  CommonsenseQA~\cite{commonsenseqa} validation set, a multiple-choice question answer dataset requiring various types of commonsense knowledge. Correctness is measured using LLM-as-a-judge.
    \item \textbf{GPQA.} (MIT License) We use the main dataset (gpqa\_main) that contains $448$ multiple-choice questions written by experts in the fields of biology, physics, and chemistry~\cite{gpqa}. Correctness is measured using LLM-as-a-judge. 
    \item \textbf{MedQA.} (MIT License) We use the $4$-option version of MedQA~\cite{medqa}, consistent with its original assessment and usage by technical reports of frontier medical LLMs~\cite{medgemini,medgemma}. Correctness is measured using LLM-as-a-judge.
    \item \textbf{MBPP.} (CC BY 4.0 License) We use 500 problems from the test split of the full version of MBPP~\cite{mbpp}, a benchmark of entry-level Python programming problems. Correctness is measured using code execution. 
\end{enumerate}

\subsubsection{Evaluation Metrics}
\label{sec:metrics}

\begin{enumerate}[nosep,leftmargin=*]
    \item \textbf{AURC ($\downarrow$).} The \emph{area under risk-coverage curve} measures cumulative selective risk as a function of coverage with sorted prediction confidence. We refer the reader to \cref{sec:technical_background} for the formal definition. Lower values means better SP performance.
    \item \textbf{ECE ($\downarrow$).} The \emph{expected calibration error} measures the alignment between confidence and correctness likelihood: 
    \begin{equation}
        \text{ECE} = \sum\nolimits_{m=1}^{M} {|B_m|}/{N} \cdot \left| \text{acc}(B_m) - \text{conf}(B_m) \right|
    \end{equation}
    where $M$ is the number of bins, $B_m$ is the set of samples in bin $m$, and $N$ is the total number of samples. We use $M=10$ bins.
    \item \textbf{Accuracy@$k\%$ ($\uparrow$).} Accuracy computed on the top $k$\% most confident predictions, measuring how well the model's confidence identifies its correct predictions. We report results at coverage levels $k \in \{10, 25, 50\}$. 
\end{enumerate}

\subsection{Mimicking Selective Prediction in Deployment}\label{app:deployment_exps_details}
\subsubsection{The MedQA Dataset}\label{app:medqa_dataset}

We use the 4-option version of MedQA~\cite{medqa}, consistent with its original assessment and usage by frontier medical LLMs~\cite{medgemini,medgemma}. Each question presents a clinical vignette followed by four answer choices ($A, B, C, D$). We partition the dataset as detailed in the following table:
\begin{table}[!htbp]
\centering
\begin{tabular}{lcc}
\toprule
\textbf{Split} & \textbf{Samples} & \textbf{Usage} \\
\midrule
Train & 10,178 & alignment training\\
Validation & 1,272 & threshold selection \\
Test & 1,273 & deployment evaluation \\
\bottomrule
\end{tabular}
\vspace{-1em}
\end{table}

The prediction correctness is checked by exact matching by the letter of the answer, i.e., among $\set{A, B, C, D}$. When the prediction does not follow the expected format, we apply robust answer extraction using multiple regex patterns: direct letter match, letter prefix (e.g., ``D. Nitrofurantoin''), ``The answer is'' prefix, and pure-option-text-without-option-letter matching. 

\subsubsection{Inference Settings}\label{app:deployment_inference}

We use the same generation settings as those of the in-domain evaluation (\cref{app:eval}): temperature set to $0$, maximum token budget of $4096$, and with the long system prompt (\cref{app:prompts}). We use vLLM~\cite{kwon2023efficient} for efficient inference. The models are loaded from HotPotQA training checkpoints. 

\subsubsection{Threshold Selection---A Key Step in Error-Controlled SP Deployment}\label{app:threshold_selection}
Given a target accuracy $\alpha^*$ and the validation set $D_{\text{val}}$, we find the threshold $\tau$ via 
\begin{equation}
    \tau = \min \left\{ t \in [0, 1] : \text{acc}(\{(x, y) \in D_{\text{val}} : s(x) \geq t\}) \geq \alpha^* \right\}, 
\end{equation}
where $s(x)$ is the confidence score for the input $x$. The corresponding coverage is computed as:
\begin{equation}
    \text{coverage}(\tau) = {|\{x \in D_{\text{val}} : s(x) \geq \tau\}|}/{|D_{\text{val}}|}. 
\end{equation}
We set $\alpha^* = 75\%$.

\subsubsection{Evaluation Metrics}\label{app:deployment_eval}
After threshold selection above, we apply the threshold directly to the test set and report the following metrics: 
\begin{itemize}[nosep,leftmargin=*]
    \item \textbf{Validation Accuracy}: the prediction accuracy on validation samples with confidence  $\ge \tau$
    \item \textbf{Validation Coverage}: the fraction of validation samples with confidence $\geq \tau$
    \item \textbf{Test Accuracy}: the prediction accuracy on test samples with confidence  $\ge \tau$
    \item \textbf{Test Coverage}: the fraction of test samples with confidence $\geq \tau$ 
\end{itemize} 

\subsubsection{Post-hoc Calibration Baseline}
\label{sec:temp-scaling-detail}
To evaluate whether post-hoc calibration can rescue poorly-calibrated confidence for SP, we
apply temperature scaling~\cite{guo2017calibration} to RLVR's confidence output. Since our setting does not generate native logits, we follow~\cite{calibrating_verbalized} which proposes to calibrate a verbalized probability vector $(p_1,\dots,p_K)$ via an invert-softmax trick: assume that the pre-softmax logits take the form $z \doteq (\log p_1+ b, \dots, \log p_K+ b)$ for an unknown constant $b$, the idea is to perform the temperature scaling $\operatorname{softmax}(z/T)$. For our single output case, we treat it as a binary classification problem on the $\{\text{correct},\text{incorrect}\}$ classes, i.e., $K =2$ so that the effective class probabilities are $(s, 1-s)$. So: 
\begin{align}
    s_T \doteq \operatorname{softmax}(z/T)_{\text{correct}}
    & = \frac{e^{(\log s + b)/T}}{e^{(\log s + b)/T} + e^{(\log (1-s) + b)/T}}
    = \frac{1}{1+ e^{\tfrac{1}{T}\log(1-s)/s}}   \\
    & = \operatorname{sigmoid}(\tfrac{1}{T}\log \tfrac{s}{1-s}), 
\end{align}
yielding our temperature scaling operator. We find an appropriate $T$ by training it over the binary calibration set $\{(s_i, y_i)\}_{i=1}^{N}$ ($s_i$:
verbalized confidences; $y_i \in \set{0, 1}$: groundtruth correctness label): 
\begin{equation}
    T^* = \argmin\nolimits_{T > 0}\; -\sum\nolimits_{i=1}^{N}
        y_i \log s_T(s_i)  + (1 - y_i) \log(1 - s_T(s_i)). 
    \label{eq:temp-fit}
\end{equation}
The optimal $T^* \approx 47.9$ on the MedQA indicates severe overconfidence of the original prediction: it squashes the
full probability domain $[0,1]$ to $\approx[0.38, 0.62]$, and the confidence
values the model actually emits, $\{0.5,\dots,1.0\}$, collapse into the even
narrower band $\approx[0.50, 0.62]$. 

It is easy to verify that $S_T$ is a monotonically increasing function of $s$. Therefore, 
\emph{it maintains the confidence ranking of the predictions---leaving the ranking-based risk-coverage curve unchanged}, as empirically confirmed in \cref{fig:medqa_aurc}. The same holds for any
monotonic scalar recalibration (e.g.\ Platt scaling). So, while confidence calibration does not necessarily improve SP performance, improving the confidence ranking is the key. 

\subsection{Experiment Details of Margin Analysis for the Lifted AURC Versus Canonical AURC}\label{app:margin_analysis}

To compare the effect of lifted AURC versus AURC as the reward on the \emph{margin}---the separation of correct and incorrect predictions in terms of confidence, we finetune the base model Qwen2.5-7B using an identical setting except for the reward function: one with lifted AURC (awarding correct predictions and penalizing incorrect predictions) and one with AURC (only penalizing incorrect predictions) on $6,000$ training samples of the HotPotQA dataset---enough for the format reward to converge for faithful parsing of confidence scores. We evaluate both models on the same $6,000$ training samples and $500$ held-out test samples. For each sample, we extract the model's predicted answer, its correctness (determined by exact match), and its verbalized confidence score. We then partition the samples into correct and incorrect predictions and estimate their respective \emph{normalized} distributions using kernel density estimation with a bandwidth of $0.15$. Finally, we report the confidence gap, i.e., the difference between the mean confidence values of the two distributions. \cref{fig:conf_dist} visualizes these distributions, with solid lines denoting correct predictions and dotted lines denoting incorrect predictions. 

\clearpage

\section{Additional Experiment Results}\label{app:more_exps_results}

\subsection{RLSR Works with Different Model Families}\label{app:additional_model}
\begin{wraptable}{r}{0.55\textwidth}
\vspace{-1.5em}
\centering
\caption{The Llama-3.1-8B results trained on HotPotQA. \colorbox{lightorange!15}{Orange} highlights the best result.}
\renewcommand{\arraystretch}{1.15}
\setlength{\tabcolsep}{3pt}
\small
\begin{tabular}{@{}lcc@{}}
\toprule
Method & In-Domain AURC$\downarrow$ & Out-of-Domain AURC$\downarrow$ \\
\midrule
BASE & 0.54 & 0.41 \\
RLVR & 0.46 & 0.42 \\
RLCR & 0.43 & 0.40 \\
\hdashline
\textbf{RLSR} & \cellcolor{lightorange!15}0.41 & \cellcolor{lightorange!15}0.38 \\
\bottomrule
\end{tabular}
\label{tab:llama_results}
\vspace{-30pt}
\end{wraptable}
We follow the same training recipe and evaluation protocol as \cref{tab:results} for the Llama-3.1-8B model. The conclusion remains the same: RLSR achieves the best AURC (lower the better) performance averaged over in-domain (HotPotQA, HotPotQA-Vanilla) and out-of-domain (SimpleQA, TriviaQA, GPQA, Math500, GSM8K, CommonsenseQA) benchmarks.

\subsection{Statistical Significance of RLSR over RLCR and RLVR}\label{app:stat_significance}

\begin{wraptable}{l}{0.5\textwidth}
\vspace{-1em}
\centering
\caption{The AURC (mean $\pm$ std) of Qwen2.5-7B trained on HotPotQA over $5$ seeds at $T=0.3$.}
\renewcommand{\arraystretch}{1.15}
\setlength{\tabcolsep}{3pt}
\small
\begin{tabular}{@{}lcc@{}}
\toprule
Method & In-Domain AURC$\downarrow$ & Out-of-Domain AURC$\downarrow$ \\
\midrule
RLVR & 0.537 $\pm$ 0.008 & 0.453 $\pm$ 0.006 \\
RLCR & 0.509 $\pm$ 0.012 & 0.443 $\pm$ 0.009 \\
\hdashline
\textbf{RLSR} & \cellcolor{lightorange!15}0.478 $\pm$ 0.006 & \cellcolor{lightorange!15}0.412 $\pm$ 0.009 \\
\bottomrule
\end{tabular}
\label{tab:stat_significance}
\vspace{-10pt}
\end{wraptable}

Our main results in \cref{tab:results} are reported using greedy decoding (temperature=$0$), where it is (almost) deterministic. To obtain variances and verify that the improvement of RLSR over the baselines is not due to sampling noise, we re-run all methods with $5$ seeds at temperature $0.3$ on HotPotQA. The non-overlapping confidence intervals in \cref{tab:stat_significance} confirm the  statistical significance of RLSR's improved performance.

\subsection{Robustness to the Choice of LLM Judge}\label{app:judge}

\begin{wraptable}{r}{0.50\textwidth}
\vspace{-15pt}
\centering
\caption{AURC of Qwen2.5-7B trained on HotPotQA, averaged over the four OOD benchmarks that use LLM-as-a-Judge (TriviaQA, SimpleQA, CommonsenseQA, GPQA), under three judges. \colorbox{lightorange!15}{Orange} highlights the best result per judge.}
\renewcommand{\arraystretch}{1.15}
\setlength{\tabcolsep}{4pt}
\small
\begin{tabular}{@{}lccc@{}}
\toprule
Judge & RLVR & RLCR & \textbf{RLSR} \\
\midrule
Llama-3.1-8B-Instruct & 0.51 & 0.50 & \cellcolor{lightorange!15}0.46 \\
Qwen2.5-32B-Instruct & 0.59 & 0.58 & \cellcolor{lightorange!15}0.56 \\
gpt-oss-20b & 0.61 & 0.60 & \cellcolor{lightorange!15}0.59 \\
\bottomrule
\end{tabular}
\label{tab:judge}
\vspace{-12pt}
\end{wraptable}

Errors of the LLM judge directly affect AURC and accuracy-at-coverage. To rule out that the advantage of RLSR stems from systematic errors of a single judge, we re-judge the generations on the four OOD benchmarks that use LLM-as-a-Judge with two additional judges from different model families and of different sizes, Qwen2.5-32B-Instruct and gpt-oss-20b~\cite{openai2025gptoss120bgptoss20bmodel}, both with greedy decoding. The absolute AURC varies between judges, but RLSR achieves the lowest AURC among all three judges (\cref{tab:judge}). 

\subsection{Evaluation on Code Generation}\label{app:mbpp}

\begin{table*}[htbp]
\centering
\caption{SP performance on MBPP, which is out-of-domain for both training settings, with the Qwen2.5-7B model. \colorbox{lightorange!15}{Orange} highlights the best result per metric within each training setting.} 
\small
\begin{tabular}{@{}lcccccc@{}}
\toprule
 & \multicolumn{3}{c}{\textbf{HotPotQA Training}} & \multicolumn{3}{c}{\textbf{BigMath Training}} \\
\cmidrule(lr){2-4} \cmidrule(l){5-7}
Method & AURC$\downarrow$ & Acc@10$\uparrow$ & Acc@25$\uparrow$ & AURC$\downarrow$ & Acc@10$\uparrow$ & Acc@25$\uparrow$ \\
\midrule
RLVR & 0.37 & 62.8\% & 62.8\% & 0.43 & 57.2\% & 57.2\% \\
RLCR & 0.40 & 60.2\% & 61.0\% & 0.36 & 70.3\% & 64.8\% \\
\hdashline
\textbf{RLSR} & \cellcolor{lightorange!15}0.34 & \cellcolor{lightorange!15}67.3\% & \cellcolor{lightorange!15}67.3\% & \cellcolor{lightorange!15}0.32 & \cellcolor{lightorange!15}77.4\% & \cellcolor{lightorange!15}72.1\% \\
\bottomrule
\end{tabular}
\label{tab:mbpp}
\end{table*}

To test generalization beyond QA and mathematical reasoning, we evaluate the final checkpoints of \cref{tab:results} on MBPP~\cite{mbpp}, a code generation benchmark used in the evaluation of Qwen2.5. Neither training set contains code, so MBPP is out-of-domain for both training settings. As shown in \cref{tab:mbpp}, RLSR attains the best AURC, Acc@10, and Acc@25 within each training setting, extending the out-of-domain advantage in \cref{tab:results} to a task type absent from both training sets. 

\subsection{Effect of Batch Size and Group Size}\label{app:batch_ablation}

\begin{table*}[htbp]
\centering
\caption{Effect of the batch size $B$ (unique prompts) and the group size $G$ (rollouts per prompt) on Qwen2.5-7B trained on HotPotQA. Rollout cost counts the rollouts generated for the same training prompts, relative to the default setting. OOD Math: GSM8K and MATH-500; OOD Free-text: CommonsenseQA, GPQA, SimpleQA, and TriviaQA. All values are AURC. \colorbox{lightorange!15}{Orange} highlights the best result per column.} 
\small
\begin{tabular}{@{}lcccccccc@{}}
\toprule
 & & & & & & \multicolumn{3}{c}{\textbf{Out-of-Domain}} \\
\cmidrule(l){7-9}
Method & $B$ & $G$ & Pool $B\times G$ & Rollout Cost & In-Domain & Math (2) & Non-Math (4) & All (6) \\
\midrule
RLVR & 48 & 32 & 1536 & 1.0$\times$ & 0.56 & 0.38 & 0.51 & 0.47 \\
RLCR & 48 & 32 & 1536 & 1.0$\times$ & 0.51 & 0.37 & 0.50 & 0.46 \\
\hdashline
\textbf{RLSR} & 48 & 32 & 1536 & 1.0$\times$ & \cellcolor{lightorange!15}0.44 & \cellcolor{lightorange!15}0.30 & 0.46 & \cellcolor{lightorange!15}0.41 \\
\textbf{RLSR} & 32 & 32 & 1024 & 1.0$\times$ & \cellcolor{lightorange!15}0.44 & 0.49 & 0.46 & 0.47 \\
\textbf{RLSR} & 16 & 32 & 512 & 1.0$\times$ & 0.45 & 0.50 & 0.47 & 0.48 \\
\textbf{RLSR} & 48 & 16 & 768 & 0.5$\times$ & 0.46 & 0.35 & 0.45 & 0.42 \\
\textbf{RLSR} & 32 & 16 & 512 & 0.5$\times$ & 0.47 & 0.39 & \cellcolor{lightorange!15}0.44 & 0.43 \\
\bottomrule
\end{tabular}
\label{tab:batch_ablation}
\end{table*}

RLSR ranks rollouts within a pooled batch of size $B\times G$ to approximate the population-level ranking in AURC, so a smaller pool introduces more finite-sample approximation noise. During training, the rollout cost depends only on $G$, while $B$ determines the number of optimization steps for a fixed amount of training data. \cref{tab:batch_ablation} shows that halving $G$, and thus the rollout cost, degrades AURC only slightly, and RLSR with $G=16$ still outperforms RLVR and RLCR at the full cost in both ID and OOD evaluation. Reducing $B$ to $G=32$ leaves the ID and free-text OOD AURC nearly unchanged, but substantially degrades the AURC on the two math OOD benchmarks. We hypothesize that this is due to domain forgetting, a common failure mode in LLM finetuning: a smaller $B$ leads to more optimization steps. The degradation is confined to the math benchmarks while ID performance is preserved, and following RLCR we train without KL regularization ($\beta=0$). We have not tested whether RLVR and RLCR show similar behavior at smaller batch sizes. 

\subsection{Comparison with Supervised Finetuning (SFT) Baseline}\label{app:sft_baselines}

\begin{wraptable}{r}{0.4\textwidth}
\vspace{-12pt}
\centering
\caption{Non-RL baseline comparison on HotPotQA with Qwen2.5-7B. SFT uses self-consistency (SC) as selector; RL methods use verbalized confidence (VC). \colorbox{lightorange!15}{Orange} highlights the best result.}
\renewcommand{\arraystretch}{1.15}
\setlength{\tabcolsep}{4pt}
\small
\begin{tabular}{@{}llc@{}}
\toprule
Method & Selector & ID AURC$\downarrow$ \\
\midrule
Base & VC & 0.60 \\
SFT-Answer-Only & SC (K=8) & 0.51 \\
\midrule
RLVR & VC & 0.56 \\
RLCR & VC & 0.51 \\
\hdashline
\textbf{RLSR} & VC & \cellcolor{lightorange!15}0.44 \\
\bottomrule
\end{tabular}
\label{tab:sft_baseline}
\vspace{-25pt}
\end{wraptable}

To isolate the effect of RL training, we fine-tune Qwen2.5-7B on (question, answer) pairs from HotPotQA without chain-of-thought or confidence output because there is no groundtruth chain-of-thought and confidence. Since the SFT model cannot output verbalized confidence (VC), we use self-consistency (SC) with temperature=$0.7$ and samples $K=8$ as a post-hoc confidence selector, where the fraction of completion agreement serves as the confidence score. We observe that SFT + SC ($0.51$) matches RLCR and outperforms RLVR, confirming that SC is a strong post-hoc selector. However, RLSR still outperforms it by $0.07$ in AURC with a single inference pass, demonstrating the advantage of end-to-end RL training over supervised alternatives.

\subsection{RLSR is Agnostic to the Choice of Confidence Selector}\label{app:conf_ablation}

\begin{table*}[htbp]
\centering
\caption{Matched-selector comparison of SP performance on two in-domain and six out-of-domain evaluation datasets with the Qwen2.5-7B model trained on HotPotQA. Each block fixes the selector and varies the alignment method. RLCR and RLSR use the selector in both training and evaluation; RLVR uses it at evaluation only. \colorbox{lightorange!15}{Orange} highlights the best result per metric within each selector.} 
\small
\begin{tabular}{@{}lcccccc@{}}
\toprule
 & \multicolumn{3}{c}{\textbf{In-Domain}} & \multicolumn{3}{c}{\textbf{Out-of-Domain}} \\
\cmidrule(lr){2-4} \cmidrule(l){5-7}
Method & AURC$\downarrow$ & Acc@10$\uparrow$ & Acc@25$\uparrow$ & AURC$\downarrow$ & Acc@10$\uparrow$ & Acc@25$\uparrow$ \\
\midrule
\multicolumn{7}{l}{\textit{Verbalized Confidence (VC)}} \\
\midrule
RLVR & 0.56 & 45.5\% & 42.0\% & 0.47 & 52.6\% & 52.9\% \\
RLCR & 0.51 & 53.5\% & 49.0\% & 0.46 & 53.5\% & 54.3\% \\
\hdashline
\textbf{RLSR} & \cellcolor{lightorange!15}0.44 & \cellcolor{lightorange!15}66.0\% & \cellcolor{lightorange!15}63.2\% & \cellcolor{lightorange!15}0.41 & \cellcolor{lightorange!15}62.0\% & \cellcolor{lightorange!15}60.7\% \\
\midrule
\multicolumn{7}{l}{\textit{Length-Normalized Log-Probability (Log-prob)}} \\
\midrule
RLVR & 0.55 & 56.0\% & 48.0\% & 0.48 & 49.7\% & 51.4\% \\
RLCR & \cellcolor{lightorange!15}0.47 & 62.5\% & 58.2\% & 0.49 & \cellcolor{lightorange!15}58.6\% & \cellcolor{lightorange!15}56.2\% \\
\hdashline
\textbf{RLSR} & \cellcolor{lightorange!15}0.47 & \cellcolor{lightorange!15}63.5\% & \cellcolor{lightorange!15}60.2\% & \cellcolor{lightorange!15}0.46 & 55.9\% & 56.0\% \\
\midrule
\multicolumn{7}{l}{\textit{Self-Consistency (SC, $K{=}8$)}} \\
\midrule
RLVR & 0.37 & 78.5\% & \cellcolor{lightorange!15}73.4\% & 0.38 & 71.8\% & 68.4\% \\
RLCR & 0.42 & 69.5\% & 68.4\% & 0.45 & 67.8\% & 60.9\% \\
\hdashline
\textbf{RLSR} & \cellcolor{lightorange!15}0.35 & \cellcolor{lightorange!15}82.0\% & \cellcolor{lightorange!15}73.4\% & \cellcolor{lightorange!15}0.37 & \cellcolor{lightorange!15}73.6\% & \cellcolor{lightorange!15}69.2\% \\
\bottomrule
\end{tabular}
\label{tab:selector_ablation}
\end{table*}

As the community has not converged to a standard for LLM confidence quantification~\cite{survey_confidence,survey_uncertainty,selectivepredictiongap}, we use verbalized confidence (VC) as the default scorer throughout the paper for its simplicity and popularity. The RLSR reward in \cref{eq:final_reward} depends on the confidence score $s_i$ only through its rank $r_i$, so any confidence score that ranks rollouts can replace VC. To test whether the advantage of RLSR is unique to VC, we retrain RLSR and RLCR on HotPotQA with Qwen2.5-7B under two additional scorers, keeping the training hyperparameters of \cref{app:training_hyperparams} and removing the request for a VC from the system prompt: (1) length-normalized log-probability (Log-prob) of each rollout; and (2) self-consistency (SC), where the score of a rollout is the fraction of the $G=32$ rollouts for the same prompt that share its answer. Rollouts with tied scores receive the average of their weights $\widehat{\alpha}_i$. For RLCR, the Brier term uses the exponentiated Log-prob or the SC agreement fraction as $s$. Since the RLVR reward does not involve confidence, we do not retrain RLVR and instead evaluate its checkpoint from \cref{tab:results} with each selector. In evaluation, the SC confidence of a query is the fraction of $K=8$ responses (one greedy, seven sampled in $T=0.7$) that agree with the greedy answer, which serves as a prediction.

\cref{tab:selector_ablation} shows that RLSR achieves the best or tied-best AURC under all three scorers in both ID and OOD evaluation. The margin is the largest with VC. With Log-prob, RLSR ties RLCR in ID AURC and trails it on OOD Acc@10 and Acc@25. With SC, RLVR is close to RLSR and ties it on ID Acc@25. RLSR with SC achieves the lowest overall AURC, at the cost of $K=8$ forward passes per query at evaluation. We keep VC as default for three reasons. First, VC requires a single forward pass, whereas SC multiplies the inference cost by $K$. Second, VC is part of the sampled response, so the policy gradient trains the answer and its confidence jointly, without the additional architectural components required by the learned scorers~\cite{aspire,selectllm}. Third, VC follows the protocol of RLCR~\cite{rlcr}, which makes the comparison with RLCR direct, and produces a confidence value that can be reported to users.

\subsection{Per-Benchmark Performance}

\cref{tab:results} reports the average performance on the 8 benchmark datasets. Here, we include the performance per-benchmark for Qwen2.5-7B trained on HotPotQA (in-domain in \cref{tab:hotpot-id} and out-of-domain in \cref{tab:hotpot-ood}) and BigMath (in-domain in \cref{tab:math-id} and out-of-domain in \cref{tab:math-ood}) for completeness. 

\begin{table}[!htbp]
\centering
\caption{The in-domain results of Qwen2.5-7B trained on HotPotQA.}
\renewcommand{\arraystretch}{1.05}
\begin{tabular}{@{}llccccc@{}}
\toprule
Dataset & Method & AURC$\downarrow$ & ECE$\downarrow$ & Acc@10$\uparrow$ & Acc@25$\uparrow$ & Acc@50$\uparrow$ \\
\midrule
HotPotQA & BASE & 0.64 & 0.60 & 44.0 & 40.0 & 34.8 \\
 & RLVR & 0.63 & 0.62 & 38.0 & 36.0 & 34.4 \\
 & RLCR & 0.55 & \cellcolor{lightorange!15}0.03 & 48.0 & 46.4 & 47.2 \\
 & \textbf{RLSR} & \cellcolor{lightorange!15}0.44 & 0.20 & \cellcolor{lightorange!15}76.0 & \cellcolor{lightorange!15}68.0 & \cellcolor{lightorange!15}55.2 \\
\hdashline
HotPotQA-Vanilla & BASE & 0.57 & 0.54 & 48.0 & 41.2 & 40.8 \\
 & RLVR & 0.50 & 0.51 & 53.0 & 48.0 & 49.2 \\
 & RLCR & 0.46 & \cellcolor{lightorange!15}0.07 & \cellcolor{lightorange!15}59.0 & 51.6 & 53.6 \\
 & \textbf{RLSR} & \cellcolor{lightorange!15}0.44 & 0.27 & 56.0 & \cellcolor{lightorange!15}58.4 & \cellcolor{lightorange!15}59.2 \\
\bottomrule
\end{tabular}
\label{tab:hotpot-id}
\end{table}

\begin{table}[htbp]
\centering
\caption{The out-of-domain results of Qwen2.5-7B trained on HotPotQA.}
\renewcommand{\arraystretch}{1.05}
\begin{tabular}{@{}llccccc@{}}
\toprule
Dataset & Method & AURC$\downarrow$ & ECE$\downarrow$ & Acc@10$\uparrow$ & Acc@25$\uparrow$ & Acc@50$\uparrow$ \\
\midrule
GSM8K & BASE & 0.24 & 0.22 & 78.8 & 78.8 & 74.2 \\
 & RLVR & 0.21 & 0.20 & 78.8 & 80.0 & 79.2 \\
 & RLCR & 0.17 & 0.33 & 80.3 & 83.6 & 84.1 \\
 & \textbf{RLSR} & \cellcolor{lightorange!15}0.16 & \cellcolor{lightorange!15}0.11 & \cellcolor{lightorange!15}81.8 & \cellcolor{lightorange!15}85.8 & \cellcolor{lightorange!15}87.9 \\
\hdashline
MATH-500 & BASE & 0.51 & 0.50 & 50.0 & 51.2 & 46.4 \\
 & RLVR & 0.56 & 0.55 & 42.0 & 43.2 & 42.4 \\
 & RLCR & 0.56 & \cellcolor{lightorange!15}0.06 & 42.0 & 45.6 & 46.8 \\
 & \textbf{RLSR} & \cellcolor{lightorange!15}0.45 & 0.32 & \cellcolor{lightorange!15}66.0 & \cellcolor{lightorange!15}57.6 & \cellcolor{lightorange!15}53.2 \\
\hdashline
CommonsenseQA & BASE & \cellcolor{lightorange!15}0.08 & \cellcolor{lightorange!15}0.01 & 92.7 & \cellcolor{lightorange!15}92.8 & \cellcolor{lightorange!15}92.5 \\
 & RLVR & 0.13 & 0.11 & 86.2 & 86.3 & 88.4 \\
 & RLCR & 0.08 & 0.40 & \cellcolor{lightorange!15}95.1 & 92.5 & 91.7 \\
 & \textbf{RLSR} & 0.09 & 0.14 & 94.3 & 92.8 & 90.8 \\
\hdashline
GPQA & BASE & \cellcolor{lightorange!15}0.55 & 0.48 & \cellcolor{lightorange!15}53.3 & \cellcolor{lightorange!15}49.1 & \cellcolor{lightorange!15}44.2 \\
 & RLVR & 0.60 & 0.60 & 44.4 & 39.3 & 39.7 \\
 & RLCR & 0.66 & \cellcolor{lightorange!15}0.15 & 24.4 & 32.1 & 37.9 \\
 & \textbf{RLSR} & 0.58 & 0.35 & 46.7 & 45.5 & 42.4 \\
\hdashline
SimpleQA & BASE & 0.86 & 0.81 & \cellcolor{lightorange!15}14.5 & 13.8 & 13.6 \\
 & RLVR & 0.89 & 0.89 & 9.7 & 11.7 & 11.1 \\
 & RLCR & 0.88 & \cellcolor{lightorange!15}0.36 & 11.8 & 11.4 & 11.5 \\
 & \textbf{RLSR} & \cellcolor{lightorange!15}0.85 & 0.50 & 12.9 & \cellcolor{lightorange!15}14.1 & \cellcolor{lightorange!15}14.8 \\
\hdashline
TriviaQA & BASE & 0.40 & 0.37 & 65.0 & 58.6 & 58.6 \\
 & RLVR & 0.42 & 0.42 & 54.5 & 56.8 & 57.8 \\
 & RLCR & 0.38 & \cellcolor{lightorange!15}0.09 & 67.5 & 60.8 & 60.0 \\
 & \textbf{RLSR} & \cellcolor{lightorange!15}0.34 & 0.27 & \cellcolor{lightorange!15}70.5 & \cellcolor{lightorange!15}68.4 & \cellcolor{lightorange!15}67.8 \\
\bottomrule
\end{tabular}
\label{tab:hotpot-ood}
\end{table}
\begin{table}[!htbp]
\centering
\caption{The in-domain results of Qwen2.5-7B trained on BigMath.}
\renewcommand{\arraystretch}{1.05}
\begin{tabular}{@{}llccccc@{}}
\toprule
Dataset & Method & AURC$\downarrow$ & ECE$\downarrow$ & Acc@10$\uparrow$ & Acc@25$\uparrow$ & Acc@50$\uparrow$ \\
\midrule
BigMath & BASE & 0.49 & 0.46 & 48.0 & 54.4 & 53.0 \\
 & RLVR & 0.37 & 0.32 & 56.0 & 63.2 & 64.6 \\
 & RLCR & 0.32 & \cellcolor{lightorange!15}0.07 & 69.0 & 69.6 & 68.0 \\
 & \textbf{RLSR} & \cellcolor{lightorange!15}0.24 & 0.19 & \cellcolor{lightorange!15}79.0 & \cellcolor{lightorange!15}80.8 & \cellcolor{lightorange!15}75.2 \\
\hdashline
GSM8K & BASE & 0.24 & 0.22 & 78.8 & 78.8 & 74.2 \\
 & RLVR & 0.12 & 0.10 & 88.6 & 87.3 & \cellcolor{lightorange!15}88.2 \\
 & RLCR & 0.13 & 0.27 & 87.1 & 86.4 & 86.5 \\
 & \textbf{RLSR} & \cellcolor{lightorange!15}0.12 & \cellcolor{lightorange!15}0.07 & \cellcolor{lightorange!15}89.4 & \cellcolor{lightorange!15}90.3 & 88.0 \\
\hdashline
MATH-500 & BASE & 0.51 & 0.50 & 50.0 & 51.2 & 46.4 \\
 & RLVR & 0.36 & 0.34 & 72.0 & 61.6 & 61.6 \\
 & RLCR & \cellcolor{lightorange!15}0.33 & \cellcolor{lightorange!15}0.03 & \cellcolor{lightorange!15}74.0 & \cellcolor{lightorange!15}70.4 & \cellcolor{lightorange!15}66.4 \\
 & \textbf{RLSR} & 0.36 & 0.29 & 68.0 & 64.0 & 66.0 \\
\bottomrule
\end{tabular}
\label{tab:math-id}
\end{table}

\begin{table}[htbp]
\centering
\caption{The out-of-domain results of Qwen2.5-7B trained on BigMath.}
\renewcommand{\arraystretch}{1.05}
\begin{tabular}{@{}llccccc@{}}
\toprule
Dataset & Method & AURC$\downarrow$ & ECE$\downarrow$ & Acc@10$\uparrow$ & Acc@25$\uparrow$ & Acc@50$\uparrow$ \\
\midrule
HotPotQA-Vanilla & BASE & 0.57 & 0.54 & 48.0 & 41.2 & 40.8 \\
 & RLVR & 0.53 & 0.53 & 51.0 & 44.8 & 46.6 \\
 & RLCR & \cellcolor{lightorange!15}0.47 & \cellcolor{lightorange!15}0.12 & \cellcolor{lightorange!15}63.0 & 55.2 & \cellcolor{lightorange!15}53.2 \\
 & \textbf{RLSR} & 0.49 & 0.36 & 55.0 & \cellcolor{lightorange!15}55.6 & 51.8 \\
\hdashline
CommonsenseQA & BASE & \cellcolor{lightorange!15}0.08 & \cellcolor{lightorange!15}0.01 & \cellcolor{lightorange!15}92.7 & \cellcolor{lightorange!15}92.8 & 92.5 \\
 & RLVR & 0.15 & 0.07 & 83.7 & 83.7 & 84.6 \\
 & RLCR & 0.11 & 0.29 & 90.2 & 89.9 & 88.9 \\
 & \textbf{RLSR} & 0.09 & 0.10 & 91.1 & 91.8 & \cellcolor{lightorange!15}92.6 \\
\hdashline
GPQA & BASE & 0.55 & 0.48 & \cellcolor{lightorange!15}53.3 & \cellcolor{lightorange!15}49.1 & 44.2 \\
 & RLVR & 0.59 & 0.59 & 42.2 & 37.5 & 40.2 \\
 & RLCR & 0.57 & \cellcolor{lightorange!15}0.18 & 48.9 & 45.5 & 39.7 \\
 & \textbf{RLSR} & \cellcolor{lightorange!15}0.53 & 0.35 & 53.3 & 48.2 & \cellcolor{lightorange!15}45.1 \\
\hdashline
SimpleQA & BASE & \cellcolor{lightorange!15}0.86 & 0.81 & \cellcolor{lightorange!15}14.5 & \cellcolor{lightorange!15}13.8 & \cellcolor{lightorange!15}13.6 \\
 & RLVR & 0.89 & 0.86 & 10.9 & 10.6 & 10.4 \\
 & RLCR & 0.88 & \cellcolor{lightorange!15}0.44 & 12.9 & 12.2 & 13.1 \\
 & \textbf{RLSR} & 0.87 & 0.65 & 14.5 & 12.5 & 12.4 \\
\hdashline
TriviaQA & BASE & 0.40 & 0.37 & 65.0 & 58.6 & 58.6 \\
 & RLVR & 0.50 & 0.40 & 44.0 & 43.0 & 52.0 \\
 & RLCR & 0.33 & \cellcolor{lightorange!15}0.04 & 71.5 & 66.4 & 65.6 \\
 & \textbf{RLSR} & \cellcolor{lightorange!15}0.30 & 0.22 & \cellcolor{lightorange!15}78.5 & \cellcolor{lightorange!15}73.6 & \cellcolor{lightorange!15}69.1 \\
\bottomrule
\end{tabular}
\label{tab:math-ood}
\end{table}

\end{document}